\documentclass{article}

\PassOptionsToPackage{numbers, compress}{natbib}
\usepackage[preprint]{neurips_2026}
\usepackage{graphicx}
\usepackage[utf8]{inputenc} 
\usepackage[T1]{fontenc}    
\usepackage{hyperref}       
\usepackage{url}            
\usepackage{booktabs}       
\usepackage{amsfonts}       
\usepackage{nicefrac}       
\usepackage{microtype}      
\usepackage{xcolor}         
\usepackage{amsmath}
\usepackage{multirow}
\usepackage[normalem]{ulem}
\usepackage{CJKutf8}
\usepackage{wrapfig}
\usepackage{enumitem}
\usepackage{subcaption} %
\usepackage{amsmath,amssymb}
\usepackage{amsthm}
\title{Adversarial Error Correction for Visual Autoregressive Generation}

\author{%
  Ligong Bi$^{1 2}$ \quad
  Tao Huang$^{1}$\thanks{Corresponding author.} \quad
  Jianyuan Guo$^{2}$ \quad
  Chang Xu$^{3}$ \\[1.5ex]
  $^{1}$Shanghai Jiao Tong University \quad
  $^{2}$City University of Hong Kong \quad
  $^{3}$The University of Sydney \\[1.5ex]
  \texttt{\small bijiw515@gmail.com, t.huang@sjtu.edu.cn, jianyguo@cityu.edu.hk, c.xu@sydney.edu.au}
}

\begin{document}

\maketitle

\begin{abstract}
Visual Autoregressive (VAR) models have emerged as a powerful paradigm for image synthesis by performing hierarchical next-scale prediction. However, VAR models are inherently prone to cascading error propagation, where subtle coarse-scale mispredictions are amplified across the hierarchy, ultimately distorting the final synthesis.
To mitigate this, we propose AID-VAR, a plug-and-play framework that enhances pre-trained VARs through Adversarially Injected Diagnosis. Instead of a standard passive generation, AID-VAR introduces a proactive error-correction mechanism inspired by the adversarial feedback in GANs. We deploy a discriminator to diagnose fidelity gaps at each scale transition, coupled with a lightweight guidance injector. This module operates as a non-invasive adapter that refines the feature manifold of a frozen VAR backbone, effectively steering the generation toward the distribution of real images without destabilizing the pre-trained latent space.
Furthermore, to rigorously evaluate this cross-scale progression, we introduce the Inter-Scale Consistency Score (ISCS), a novel metric that quantifies the fidelity and structural alignment between consecutive resolution scales.
Experimental results across various backbones demonstrate that AID-VAR delivers sharper textural details and fewer structural distortions with negligible overhead. For instance, AID-VAR-d20 achieves a 16\% improvement in FID with only a 3\% increase in parameters. These results establish AID-VAR as a highly efficient and scalable pathway for upgrading large-scale VAR generators—enhancing global coherence and local detail without altering training data, base architectures, or sampling schedules. Code is available at \url{https://github.com/bijiw515/AID-VAR}.
\end{abstract}

\section{Introduction}
\label{sec:intro}

\begin{figure}[htbp]
\centering
\includegraphics[width=0.9\linewidth]{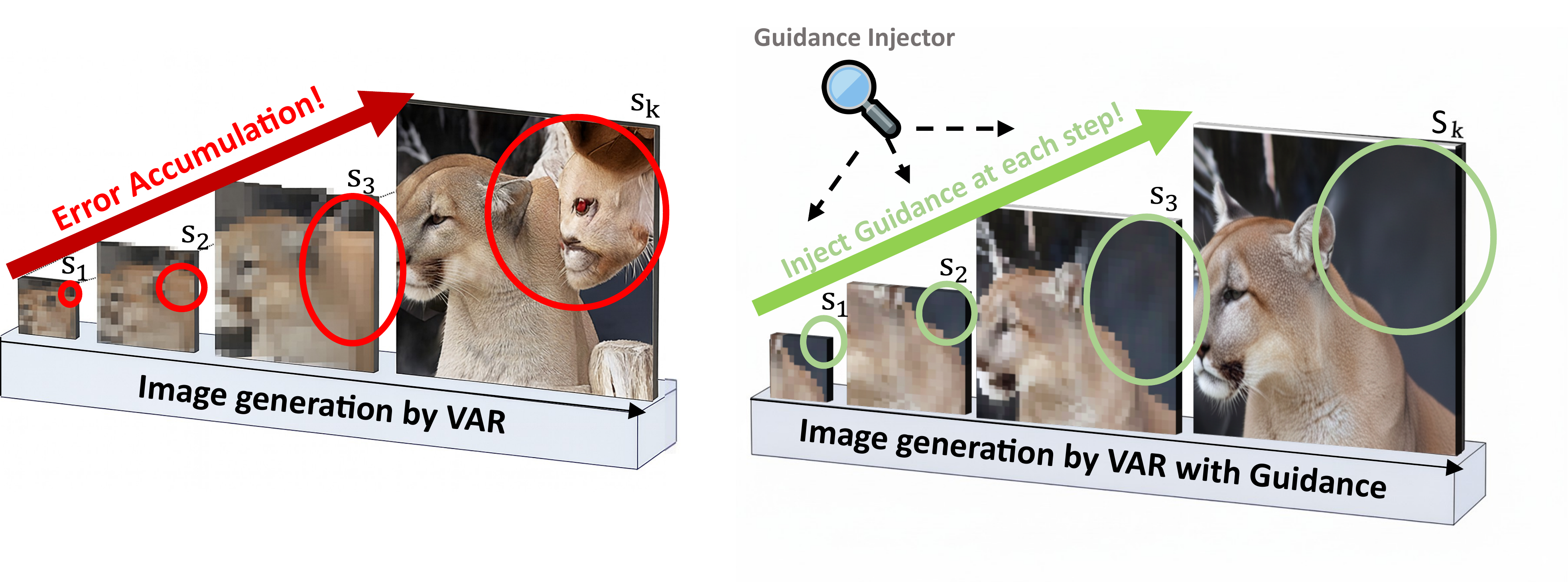}
\vspace{-16pt}
\caption{\small{
Conceptual illustration of our motivation. 
\textbf{Left:} The standard VAR model's coarse-to-fine generation is prone to error accumulation, 
where initial inaccuracies (circled in red) are magnified across scales, leading to severe structural degradation. 
\textbf{Right:} Our AID-VAR framework introduces a guidance injector at each step to anticipate and correct these errors, 
ensuring a globally coherent and plausible final image.
}}
\vspace{-16pt}
\label{fig:motivation}
\end{figure}

\begin{wrapfigure}{r}{0.56\textwidth}
    \centering
    \vspace{-12pt}
    \includegraphics[width=0.56\textwidth]{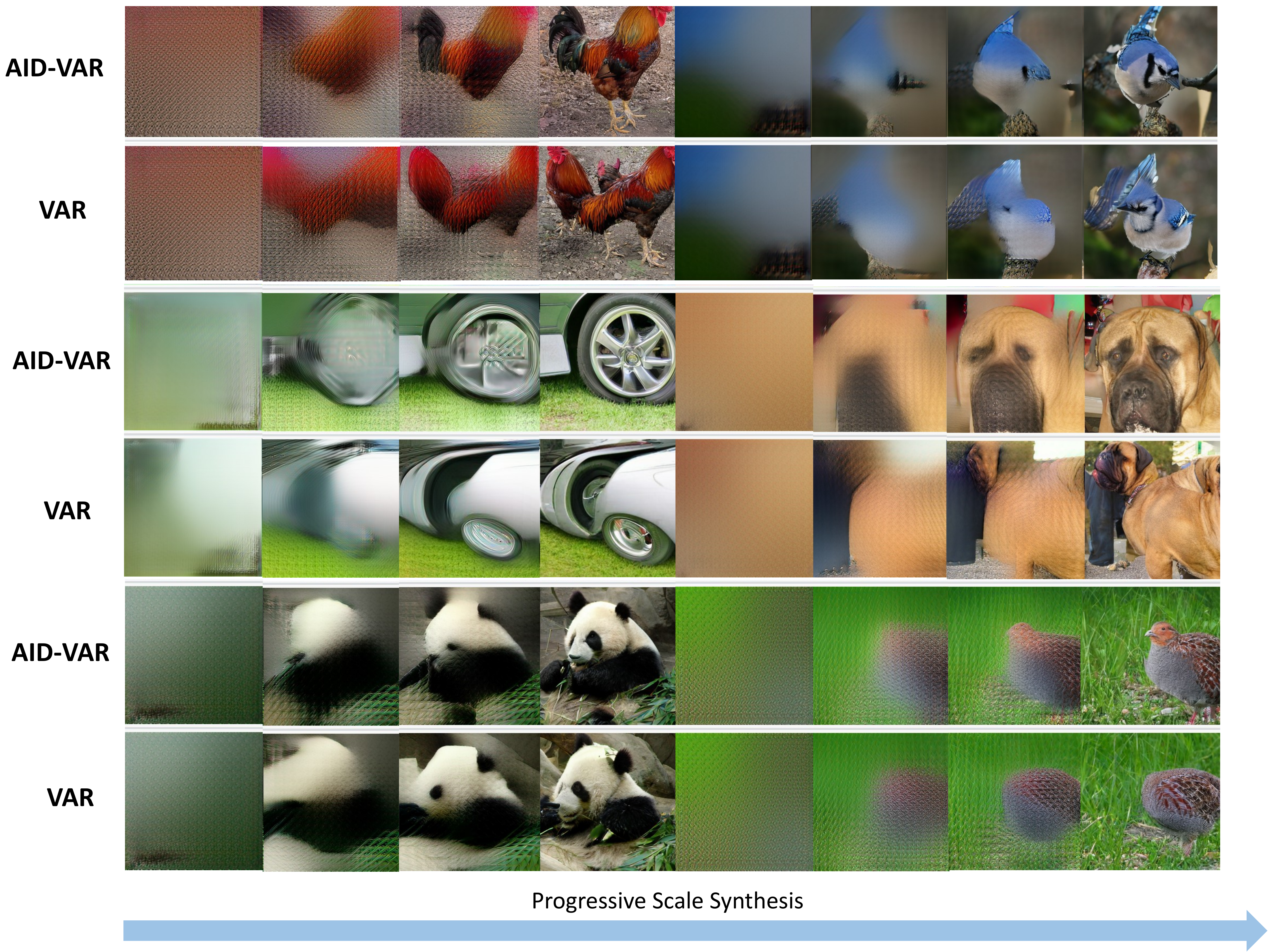}
    \vspace{-16pt}
    \caption{\small{Qualitative comparison between our AID-VAR and the standard VAR across generation scales. The comparison demonstrates that AID-VAR better preserves visual consistency and reduces error accumulation during progressive generation.}}
    \vspace{-6pt}
    \label{fig:aid_var_comparison}
\end{wrapfigure}

Autoregressive modeling has recently emerged as a competitive paradigm for high-fidelity image synthesis, offering stable training, exact likelihood, and flexible conditioning \cite{van2016pixel, van2017neural, esser2021taming, chang2022maskgit, tian2024visual}. In the visual autoregressive (VAR) \cite{tian2024visual} setting, an image is synthesized across a hierarchy of spatial scales; each stage conditions on previously generated content and predicts the next chunk of visual information. Despite their strong generative capacity, VARs suffer from a characteristic failure mode: local prediction errors accumulate across scales, compounding into detail loss, texture drift, and small-region distortions that undermine global coherence, as illustrated in Figure~\ref{fig:motivation} and Figure~\ref{fig:aid_var_comparison}. These issues stem from exposure bias \cite{bengio2015scheduled, ranzato2015sequence, schmidt2019generalization, he2019quantifying} and imperfect cross-scale transitions: once a local mistake enters the chain, subsequent decoders amplify rather than correct it \cite{parthipan2024defining, pasini2024continuous, lee2022autoregressive, tian2024visual, cheng2025tensorar, lin2024spot, yue2025understand}.

\emph{We argue that mitigating these cross-scale errors requires diagnostic feedback at each scale transition, capable of identifying ``off-manifold'' artifacts and nudging the generation back toward the natural image distribution.} Adversarial learning \cite{goodfellow2014generative, brock2018large, karras2019style, karras2021alias} provides precisely such a signal: a discriminator trained on real \textit{v.s.} generated images excels at detecting fine-grained visual pathologies (\textit{e.g.}, broken micro-geometry, aliasing, texture misalignment). However, naïvely adversarially fine-tuning the base VAR is often impractical: it introduces training instability, alters a large number of parameters, and demands substantial compute to retain the original model's strengths.

This paper introduces AID-VAR (Adversarially Injected Diagnosis for VAR), a lightweight, plug-and-play guidance module that upgrades a pre-trained, frozen VAR with adversarial feedback, without modifying training data, the base architecture, or the sampling schedule. AID-VAR first trains a discriminator to detect visual errors and leverages it to supervise a small guidance injector attached to the VAR. During generation, the injector conditions on previously generated scales and injects a spatial guidance signal that steers the next scale’s prediction toward the natural-image manifold identified by the discriminator. Crucially, the base VAR remains frozen; only the tiny injector and the discriminator heads are trained, yielding stability, efficiency, and easy implementation across backbones.

To implement the above error-fixing mechanism via adversarial guidance, a key question is where the discriminator should operate. While the VAR's native latent space, where multi-scale tokens reside, seems natural, our experiments show that discriminators trained directly in this compressed space suffer from rapid collapse and weak gradients. Latent representations, though semantically rich, lack the explicit spatial and textural details needed to discriminate subtle visual artifacts.
We therefore propose a differentiable pipeline that operates in RGB pixel space rather than latent space.
Specifically, we employ a projected discriminator~\cite{sauer2021projected} that leverages a fixed, pretrained image encoder to extract robust perceptual features and learns only shallow classification heads on top. To bridge the gap between RGB-space supervision and latent-space optimization, we introduce soft-label based VQ-VAE decoding. Unlike standard hard quantization that blocks gradients, our soft decoding computes continuous weighted combinations of codebook vectors, providing stable gradients from pixel space back through the decoder to the injector. This maintains end-to-end differentiability while preserving the discrete structure of the VAR.
Finally, to prevent the discriminator from overfitting to stale artifacts, we adopt a dynamic training refresh strategy. At each training iteration: (1) the frozen VAR paired with the current injector generates refined images via the soft-decoding path; (2) the discriminator is updated on real images versus these current refined samples; (3) the injector is updated using the discriminator's adversarial signal. This on-the-fly refresh stabilizes the min-max game, reduces overfitting to outdated generation patterns, and improves sample efficiency.

We further observe that hierarchical autoregressive generation calls for evaluation beyond final-image realism. Standard image-level metrics, including FID~\cite{heusel2017gans}, IS~\cite{salimans2016improved}, and precision/recall~\cite{kynkaanniemi2019improved}, mainly measure the quality of completed images, but do not reveal whether each refinement stage remains faithful to its coarse-scale context. To fill this gap, we introduce the Inter-Scale Consistency Score (ISCS), which measures the fidelity of transitions between consecutive scales. ISCS serves as a targeted probe for exposure bias across the hierarchy: higher ISCS indicates that fine-scale refinements remain consistent with coarse-scale structure, aligning with perceptual coherence. Evaluated under ISCS, baseline VARs score low due to sensitivity to early errors and semantic drift; in contrast, our AID-VAR substantially boosts ISCS, indicating stabilized cross-scale refinement and reduced error propagation. Our main contributions are summarized as follows:
\begin{itemize}[leftmargin=*, itemsep=0pt, topsep=0pt]
    \item We propose a novel adversarial framework to address the cumulated error issue in VAR, by diagnosing and correcting cross-scale errors via a lightweight injector.
    \item We make adversarial supervision effective and stable for VARs through an RGB-space discriminator and a differentiable soft-decode route that jointly deliver informative gradients to the injector, together with a dynamically refreshed training loop that maintains up-to-date fake samples.
    \item We introduce ISCS to quantify cross-scale coherence, complementing FID/IS by directly targeting exposure-bias effects between adjacent scales.
    \item Extensive experiments on ImageNet-1K~\cite{deng2009imagenet} demonstrate that AID-VAR consistently improves multiple VAR backbones while introducing only modest additional overhead.
\end{itemize}

\section{Related Work}
\label{related_work}

\textbf{Autoregressive models.}
The application of autoregressive (AR) models to image generation, inspired by their success in NLP \citep{brown2020language, radford2018improving, devlin2019bert}, has evolved through several paradigms. Early methods focused on next-pixel prediction on raster-scanned sequences \citep{van2016pixel, van2016conditional}. A major advance was the introduction of visual tokenizers like VQ-VAE \citep{van2017neural} and VQGAN \citep{esser2021taming}, which compress image patches into discrete tokens, shifting the task to next-token prediction \citep{ramesh2021zero, yu2022scaling, yu2021vector} and significantly improving quality. While this approach achieved state-of-the-art results, it suffered from an efficiency bottleneck due to the large number of required decoding steps. The Visual Autoregressive (VAR) model \citep{tian2024visual} recently addressed this with a ``next-scale prediction'' paradigm, drastically reducing generation steps. Other related works on efficient autoregressive generation include non-sequential or parallel decoding methods \citep{chang2022maskgit, li2023mage, tschannen2024givt}. Despite this progress, all AR frameworks are inherently susceptible to error accumulation due to their sequential nature \cite{bengio2015scheduled, ranzato2015sequence, schmidt2019generalization, parthipan2024defining, he2019quantifying, pasini2024continuous, tian2024visual, cheng2025tensorar, lin2024spot}. Our work, AID-VAR, directly tackles this persistent challenge by introducing an external guidance mechanism to correct the generative process without altering the base model.

\textbf{Adversarial training for generative models.}
Adversarial training, first introduced by Generative Adversarial Networks (GANs) \citep{goodfellow2014generative}, has expanded beyond its original application of training standalone GANs \citep{radford2015unsupervised, karras2019style, brock2018large, zhang2019self}. A prominent trend in recent years has been the use of adversarial objectives to enhance and refine other classes of generative models, such as diffusion models \citep{dhariwal2021diffusion, ho2020denoising, song2020denoising, rombach2022high}. For instance, adversarial losses have been used to improve perceptual quality \citep{rombach2022high}, enable fast sampling \citep{lu2022dpm, salimans2022progressive}, and distill pre-trained models \citep{sauer2024adversarial}. Our training strategy is particularly inspired by Adversarial Diffusion Distillation (ADD) \citep{sauer2024adversarial}, which effectively combines a distillation loss with an adversarial loss to achieve highly efficient, few-step image synthesis. AID-VAR specifically adopts the efficient discriminator architecture popularized by ADD and Projected GANs \citep{sauer2021projected}: a powerful, pre-trained feature extractor is kept frozen, and only a lightweight classification head is trained on top of its features. However, our application of this principle is novel. Instead of integrating the adversarial loss into the training of the primary generative model, we leverage this highly efficient adversarial framework to exclusively train an independent, lightweight guidance module---the guidance injector. This allows us to surgically enhance a pre-existing, frozen model without incurring the costs of end-to-end adversarial training.

\begin{figure}[t]
\centering
\includegraphics[width=0.9\linewidth]{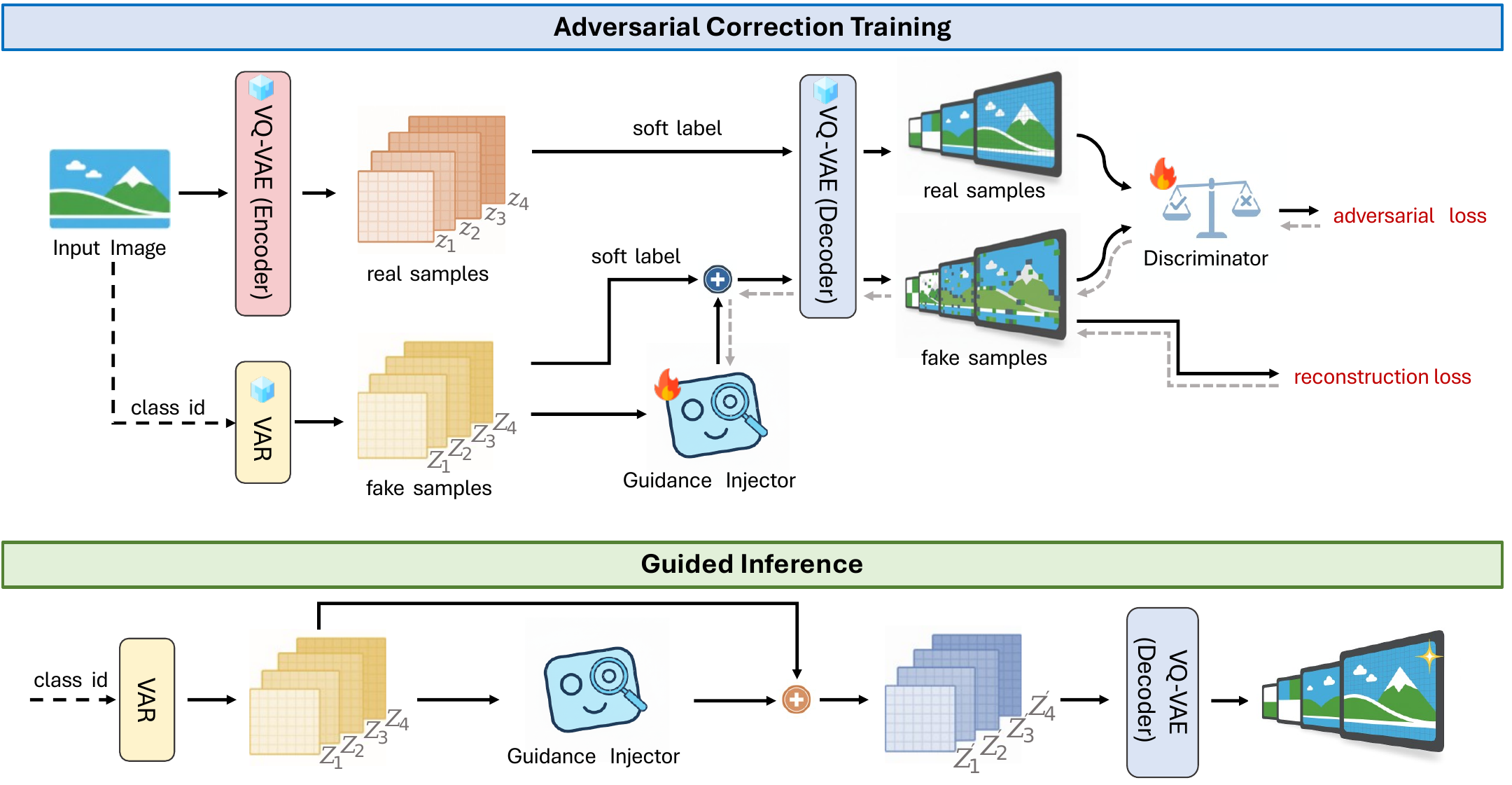} 
\vspace{-2mm}
\caption{\small{
\textbf{AID-VAR framework for guided autoregressive generation.}
(1) \textbf{Training:} We freeze the pretrained VAR and learn a discriminator to adversarially train a lightweight guidance injector, 
which produces spatial guidance maps that are injected into the VAR’s next-scale predictions via a differentiable soft-decoding path. 
(2) \textbf{Inference:} Only the guidance injector is attached to the frozen VAR to provide supplemental guidance signals, 
yielding a simple, efficient drop-in upgrade of VAR.
}}
\label{fig:framework}
\end{figure}

\section{AID-VAR for Error Diagnosis}
\label{method}
Given a frozen VAR that predicts multi-scale tokens, we aim to reduce cross-scale error accumulation without modifying its weights, data, or sampling schedule. AID-VAR introduces a lightweight \emph{guidance injector}, trained adversarially with a projected RGB-space \emph{discriminator}, to produce spatial guidance from previous scales and inject it into the next-scale prediction. The overall framework is shown in Figure~\ref{fig:framework}.

\subsection{Learning to discriminate scales} 

\textbf{Preliminary: next-scale prediction in VARs.} Visual autoregressive models synthesize an image over a coarse-to-fine hierarchy of scales. At scale $k$, the model predicts the next block of visual tokens conditioned on all previously generated tokens (and any optional external condition $c$, \textit{e.g.}, class/text). Let $S_k$ denote the predicted tokens at scale $k$, $x_k$ the corresponding hidden state, and $z_k$ the pre-softmax logits. A generic VAR defines
\begin{equation}
z_{k} = f_{\text{VAR}}(x_{k}|\{S_{<k}\}), \quad S_{k} = \arg\max \text{softmax}(z_{k}),
\label{eq:var}
\end{equation}
where each $x_k$ is conditioned on $\{S_{<k}\}$ and $c$. The process proceeds from coarse to fine and generates $K$ intermediate scales; after the finest scale, tokens are decoded to an RGB image by a VQ-VAE decoder \citep{van2017neural}.
However, this sequential generation process over discrete tokens inevitably leads to severe error accumulation, a crucial limitation we formally prove in Appendix~\ref{sec:error-accumulation}.

\textbf{Learning to discriminate intermediate scale errors.} We seek a discriminator that is maximally sensitive to fine-grained visual defects that emerge in the generated intermediate scales, yet stable enough to supervise a lightweight guidance injector without updating the frozen VAR (as theoretically analyzed in Appendix~\ref{sec:aid-var-theory}).

\begin{figure}[t] 
\centering
\includegraphics[width=0.9\linewidth]{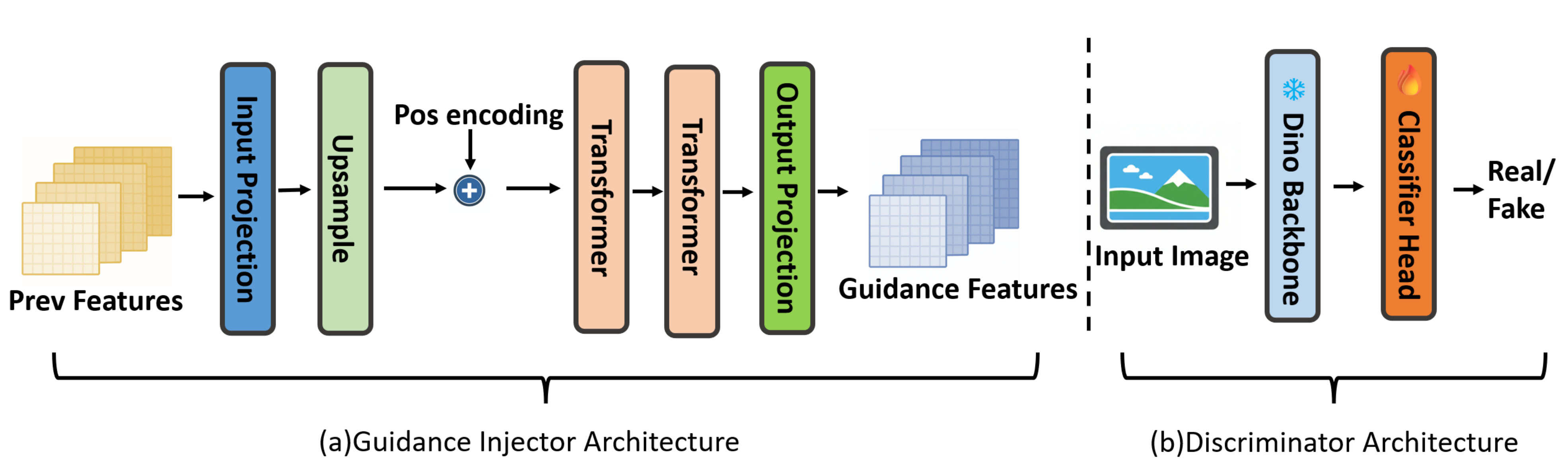}
\caption{\small{Architectures for the (a) guidance injector and (b) discriminator. The injector is a lightweight Transformer encoder that processes previous scale features. The discriminator uses a frozen DINO backbone with a trainable classification head.}}
\label{fig:architectures}
\end{figure}

Therefore, following ADD~\citep{sauer2024adversarial}, a more promising way to construct the discriminator is leveraging a pre-trained visual foundation model (\textit{e.g.}, DINO \citep{caron2021emerging}, CLIP \citep{radford2021learning}) that generalize well to the image samples. Formally, as the architecture shown in Figure~\ref{fig:architectures}(b), given an input image $I$, a frozen, pretrained vision encoder $E$ extracts features, which feed into shallow, learnable heads $h$:
\begin{equation}
    D(I;\theta) := h(E(I); \theta),
\end{equation}
where only $\theta$ is trained, and $E$ is frozen.
We empirically find this DINO-based discriminator enjoys multiple benefits compared to randomly-initialized discriminator:
\begin{itemize}[leftmargin=*, itemsep=0pt, topsep=0pt]
    \item Perceptual alignment: pretrained encoders in RGB capture natural-image statistics that are directly predictive of aliasing, texture drift, and micro-geometry breaks.
    \item Stability: discriminating in discrete/token spaces introduces distributional mismatch and early collapse, whereas RGB features are smoother and more linearly separable for real/fake cues.
    \item plug-and-play: RGB-space supervision is agnostic to tokenizer/codebook design and generalizes across VAR backbones.
\end{itemize}

\textbf{Training data.} The discriminator itself can be regarded as a binary classifier to identify real and fake samples. So the training data of discriminator is composed with real samples and fake (generated) samples: (1) Real set: We sample the image $I$ uniformly from the training dataset of VAR, then feeds it into the VAR encoder to obtain tokens at each scale $k$, and use the decoder to obtain the corresponding images $I_k$ at each scale. (2) Fake set: we run the inference of the frozen VAR to generate next-scale tokens $S_k$ at each scale $k$. We then obtain the RGB images $\hat{I}_k$ via the decoder.

\textbf{Objective.} We use hinge loss \citep{lim2017geometric} to train the discriminator on real and fake samples:
\begin{equation}
\mathcal{L}_D = \frac{1}{K}\sum_{k=1}^K \left\{ \mathbb{E}_{I_k}\big[\max(0, 1 - D(I_k;\theta))\big] + \mathbb{E}_{\hat{I}_k}\big[\max(0, 1 + D(\hat{I}_k;\theta))\big] \right\}
\label{eq:L_dis}
\end{equation}

As a result, we now have a discriminator that can identify real images and fake generated images at each scale. Next, we will discuss how we use the discriminator to adversarially force the VAR to precisely generate images that lie in the manifold of real images.

\subsection{Adversarial next-scale prediction} 

With a discriminator, the traditional adversarial approach updates the generator to fool a discriminator. Applied to VARs, this entails backpropagating through a large backbone and modifying millions (or billions) of weights, which (1) incurs substantial computational overhead; (2) destabilizes training under a min–max objective; and (3) risks irrecoverable drift in pretrained knowledge (catastrophic forgetting, degraded generalization). To avoid these pitfalls, AID-VAR is implemented as a lightweight extension to a frozen VAR: we keep \textbf{all} VAR parameters unchanged and learn only a small guidance module that nudges the next-scale prediction.

\textbf{Architecture of the guidance injector.} The guidance injector, $P_\phi$, is designed as a lightweight, spatially-aware module, as illustrated in Figure~\ref{fig:architectures}(a). It takes the feature map from the previously generated scale as input. The architecture consists of an input projection layer, an upsampling block to match the spatial dimensions of the next scale, followed by the addition of positional encodings. The core of the module is a series of two standard Transformer blocks \citep{vaswani2017attention}, which process the spatially-rich features. Finally, an output projection layer produces the guidance feature map $G_k$, which has the same dimensions as the VAR's internal hidden state for the current scale. The predicted logits $z_k$ of each scale thus becomes:
\begin{equation}
    z'_k = f_\text{VAR}(x_k + w\cdot G_k | \{S_{<k}\}),
\end{equation}
where $w$ is the guidance weight that controls the strength of the injected guidance signal $G_k$. 

\textbf{Soft-label decoding for differentiability.} The VQ-VAE path in typical VARs is discrete, in which the direct token selection $S_k=\arg\max \text{softmax}(z_{k})$ is non-differentiable. To optimize $P_\phi$ with adversarial signals in RGB space, we introduce a soft-label decoding route that preserves gradients from image space back to $z'_k$. The new decoding replaces the original discrete codebook lookup operation $H = W_{[S_k]}$ with
\begin{equation}
    H' = W\cdot\text{softmax}(z'_{k}),
\end{equation}
where $W$ is the codebook embedding matrix. This soft reconstruction enables a discriminator to act in RGB while providing stable gradients to the injector.

Note that to prevent the discriminator from exploiting reconstruction artifacts specific to the decoding route, we match its input pipeline for both real and generated samples in training data. This means both real and generated samples use soft-label decoding to obtain the images fed into the discriminator.

\textbf{Adversarial loss.} Similar to the adversarial loss in Equation (\ref{eq:L_dis}), the injector at each scale $k$ is guided with a hinge-style adversarial loss:
\begin{equation}
\mathcal{L}_{\text{adv}} = -\mathbb{E}_{\hat{I}'_k}[D(\hat{I}';\theta)],
\end{equation} 
where $\hat{I}'_k$ is the predicted image with guidance injector.
To prevent over-aggressive deviation from the VAR prior and stabilize training, we add the token-level reconstruction loss $\mathcal{L}_{\text{rec}}$ in VAR as an auxiliary objective. The overall training objective for the guidance injector $P_\phi$ thus becomes:

\begin{equation}
\mathcal{L}_P = \mathcal{L}_{\text{adv}} + \lambda_{\text{rec}}\mathcal{L}_{\text{rec}}, \quad \text{where} \quad \mathcal{L}_{\text{rec}} = \mathbb{E}_{k}[\text{CrossEntropy}(z'_k, s_k)].
\label{eq:L_P}
\end{equation}

$\mathcal{L}_{\text{rec}}$ measures the cross-entropy between the guided logits $z'_k$ and ground-truth tokens $s_k$, and $\lambda_{\text{rec}}$ is a balancing hyperparameter.

\subsection{Guided inference} 

AID-VAR's inference follows a guided autoregressive procedure. The initial scale ($k=1$) is generated using standard VAR sampling with Classifier-Free Guidance (CFG) \citep{ho2022classifier}. For all subsequent scales ($k > 1$), the guidance injector dynamically generates a guidance map $G_k$ from the previously sampled tokens $S_{k-1}$. This map is then injected into the VAR's internal state with a small, fixed weight $w$ before the current scale's tokens are sampled. This iterative, guided process continues until the final scale, after which the VQVAE decodes the complete token sequence into the final image.

\section{Inter-Scale Consistency Score (ISCS)}
\label{iscs}

The phenomenon of error accumulation in VAR models can be characterized as a scale-by-scale exposure bias \citep{ranzato2015sequence, bengio2015scheduled, schmidt2019generalization, he2019quantifying, parthipan2024defining, pasini2024continuous}, where the model is conditioned on its own, potentially flawed, outputs during inference. To specifically quantify this, we introduce the Inter-Scale Consistency Score (ISCS), a novel metric designed to measure the fidelity of the generative transitions between consecutive scales. Our core hypothesis is that a model's ability to progressively refine details and correct artifacts at finer scales is a critical indicator of its overall quality. Existing metrics like FID \citep{heusel2017gans} do not isolate this crucial aspect of refinement capability across the generative hierarchy.

To address this, we propose ISCS, which measures the Fréchet distance between the joint distributions of adjacent scales. For a large set of real and generated images, we construct two corresponding sets of joint feature vectors for each scale transition $k-1 \to k$. Using a pre-trained DINO model \citep{caron2021emerging} as a feature extractor, $\text{feat}(\cdot)$, we define the real joint distribution features $J_{\text{real}}^k$ and the generated joint distribution features $J_{\text{gen}}^k$ as:
\begin{equation}
J_{\text{real}}^k = \{[\text{feat}(r_{k-1}),~ \text{feat}(r_k)]\}, \quad
J_{\text{gen}}^k  = \{[\text{feat}(r'_{k-1}),~ \text{feat}(r'_k)]\}.
\label{eq:J_pairs}
\end{equation}

The per-scale score is the inverse of the Fréchet distance (FD) between these two sets of features:
\begin{equation}
\text{ISCS}_k = \frac{1}{\text{FD}(J_{\text{real}}^k, J_{\text{gen}}^k)}
\end{equation}
A higher score indicates a smaller distance and thus better consistency, which represents a smoother and more flawless transition from a coarser scale to a finer one. The final score is a weighted sum of the per-scale scores:
\begin{equation}
\text{ISCS} = \sum_{k=1}^{K}\alpha_k \cdot \text{ISCS}_k
\end{equation}
To emphasize late-stage generation, we use exponential weights $\alpha_k \propto 2^k$, assigning larger weights to finer scales. This biases the metric toward high-fidelity detail synthesis and artifact removal in the final refinement steps, as shown in Figure~\ref{fig:iscs_per_scale}, later stages are more informative and critical.

\begin{table}[t] 
\caption{\small{Quantitative results on the ImageNet 256$\times$256 validation set, comparing our proposed AID-VAR with its unguided counterparts and representative state-of-the-art generative models.
Metrics include Fréchet inception distance (FID), inception score (IS), precision (Pre) and recall (Rec).
`↓` indicates lower is better, while `↑` indicates higher is better.}}
\label{tab:main_results}
\begin{center}
\small
\renewcommand{\arraystretch}{0.9}
\setlength{\tabcolsep}{6pt}
\begin{tabular}{llcccccc}
\toprule
\textbf{Type} & \textbf{Model} & \textbf{Params} & \textbf{Steps}↓ & \textbf{FID}↓ & \textbf{IS}↑ & \textbf{Pre}↑ & \textbf{Rec}↑ \\
\midrule
\multirow{3}{*}{GAN} & BigGAN \citep{brock2018large} & 112M & 1 & 6.95 & 224.5 & 0.89 & 0.38 \\
& GigaGAN \citep{kang2023scaling} & 569M & 1 & 3.45 & 225.5 & 0.84 & 0.61 \\
& StyleGAN-XL \citep{sauer2022stylegan} & 166M & 1 & 2.30 & 265.1 & 0.78 & 0.53 \\
\midrule
\multirow{5}{*}{Diffusion} & ADM \citep{dhariwal2021diffusion} & 554M & 250+ & 10.94 & 101.0 & 0.69 & 0.63 \\
& LDM-4 \citep{rombach2022high} & 400M & 250+ & 3.60 & 247.7 & - & - \\
& DiT-XL/2 \citep{peebles2023scalable} & 675M & 250 & 2.27 & 278.2 & 0.83 & 0.57 \\
& L-DiT-7B \citep{alphavllm2024} & 7B & 250 & 2.28 & 316.2 & 0.83 & 0.58 \\
\midrule
\multirow{13}{*}{Autoregressive} & VQGAN \citep{esser2021taming} & 1.4B & 256 & 15.78 & 74.3 & - & - \\
& RQ-Transformer \citep{lee2022autoregressive} & 3.8B & 68 & 7.55 & 134.0 & - & - \\
& MaskGIT \citep{chang2022maskgit} & 227M & 8 & 6.18 & 182.1 & 0.80 & 0.51 \\
& MAGVIT-v2 \citep{yu2023magvit} & 307M & - & 3.65 & 200.5 & - & - \\
& MAR \citep{li2024autoregressive} & 943M & - & 2.35 & 227.8 & 0.79 & 0.62 \\
\cmidrule{2-8}
& VAR-d16 \citep{tian2024visual} & 310M & 10 & 3.55 & 274.4 & 0.84 & 0.51 \\
& AID-VAR-d16 (Ours) & 321M & 10 & \textbf{3.24} & \textbf{280.0} & \textbf{0.85} & \textbf{0.51} \\
\cmidrule{2-8}
& VAR-d20 \citep{tian2024visual} & 600M & 10 & 2.95 & 302.6 & 0.83 & 0.56 \\
& AID-VAR-d20 (Ours) & 619M & 10 & \textbf{2.54} & \textbf{309.4} & \textbf{0.83} & \textbf{0.56} \\
\cmidrule{2-8}
& VAR-d24 \citep{tian2024visual} & 1.0B & 10 & 2.33 & 312.9 & 0.82 & 0.59 \\
& AID-VAR-d24 (Ours) & 1.02B & 10 & \textbf{2.08} & \textbf{316.3} & \textbf{0.82} & \textbf{0.60} \\
\bottomrule
\end{tabular}
\end{center}
\end{table}
\section{Experiments}
\label{experiments}

\subsection{Experimental settings} 

We conduct our training and evaluation on the ImageNet-1K dataset \citep{deng2009imagenet}. All images are processed at a resolution of 256x256. The training is performed on the official training split, and all results are reported on the validation split. To ensure a fair and direct comparison, all experimental results presented are based on the same set of generated images using a fixed random seed and the other hyper parameters. This allows for a causal and related analysis between the outcomes of each experiment.

\textbf{Evaluation metrics.} We evaluate the performance of our method using standard image generation metrics: Fréchet Inception Distance (FID) \citep{heusel2017gans}, Inception Score (IS) \citep{salimans2016improved}, Precision,  Recall \citep{kynkaanniemi2019improved, sajjadi2018assessing}, and the ISCS Score that we presented.

\textbf{Implementation details.} Our experiments are built upon three pre-trained and frozen VAR models of varying sizes: VAR-d16 (310M parameters), VAR-d20 (600M), and VAR-d24 (1.0B) \citep{tian2024visual}. The trainable Guidance Injector is a lightweight Transformer with 2 layers and 8 attention heads. The discriminator is based on the StyleGAN-T architecture \citep{sauer2023stylegan}, using a frozen DINO ViT \citep{caron2021emerging} as its feature extraction backbone. The Guidance Injector and the discriminator are both trained with a learning rate of 1e-6. For the composite loss of the Guidance Injector, the reconstruction loss weight is set to $\lambda_{\text{rec}} = 0.01$. The guidance weight during the training phase is fixed at $w = 0.001$.

\begin{wrapfigure}[10]{r}{0.6\textwidth} 
    \centering
    \vspace{-26pt}
    \includegraphics[width=0.6\textwidth]{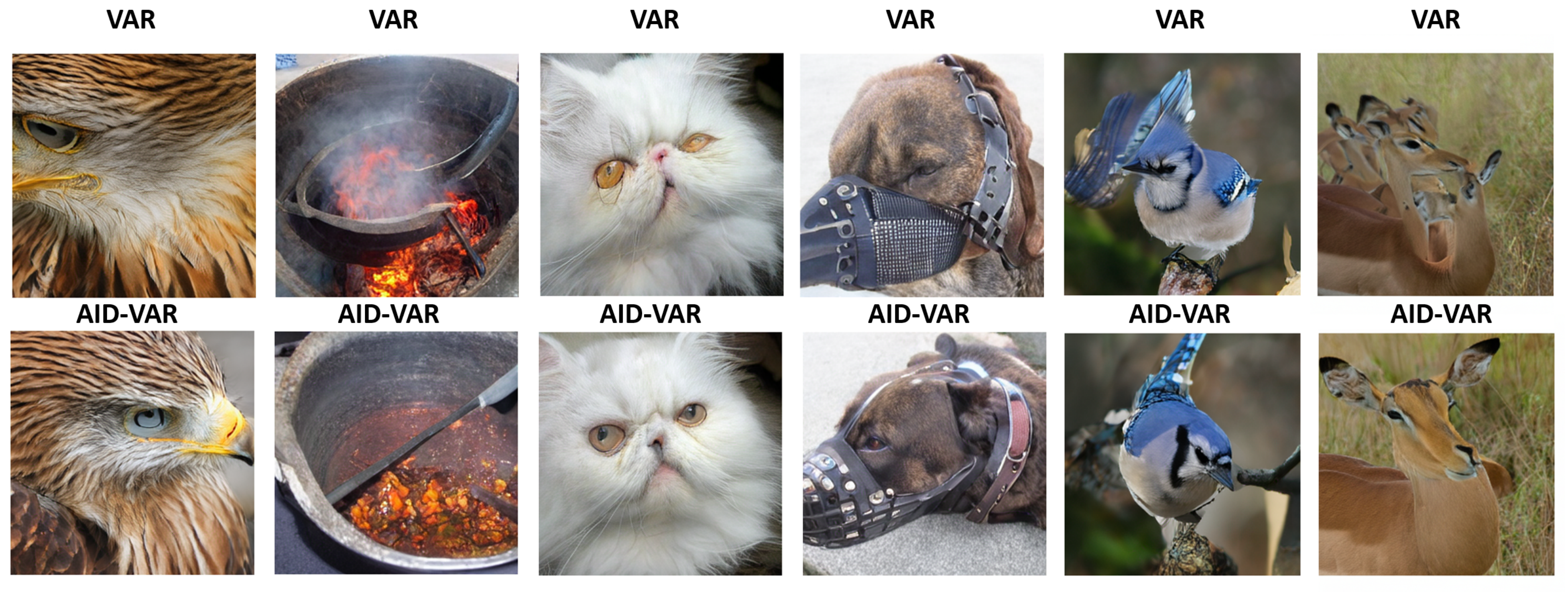}
    \vspace{-16pt}
    \caption{\small{Qualitative comparison of AID-VAR against the baseline VAR. Our method consistently corrects a wide range of structural and semantic errors present in the unguided baseline.}}
    \vspace{-8pt}
    \label{fig:qualitative_comparison}
\end{wrapfigure}

\subsection{Performance analysis} 

We first present a comprehensive performance analysis of our proposed AID-VAR framework. We evaluate our method on the ImageNet 256$\times$256 validation set and compare it against two primary sets of baselines: (1) the original, unguided VAR models of corresponding sizes, to directly measure the improvements conferred by our plug-and-play module, and (2) a wide array of state-of-the-art generative models, including leading GANs, diffusion models, and other autoregressive architectures.

The quantitative results are summarized in Table~\ref{tab:main_results}. Our findings clearly demonstrate the efficacy of the AID-VAR framework. For each model size (d16, d20, d24), our guided approach consistently outperforms its unguided VAR counterpart, achieving a notable reduction in FID while maintaining or improving other metrics. For instance, AID-VAR-d16 improves the FID score from 3.55 to 3.24. This enhancement is achieved with a negligible increase in parameters (approx. 11M), underscoring the efficiency of the Guidance Injector. When compared to the broader landscape of generative models, AID-VAR demonstrates highly competitive performance against top-tier models like MAR and DiT-XL/2, validating our approach as a practical and effective method for post-hoc enhancement of large-scale generative models.

\begin{wrapfigure}{r}{0.7\textwidth}
\centering
\vspace{-10pt}
\includegraphics[width=\linewidth]{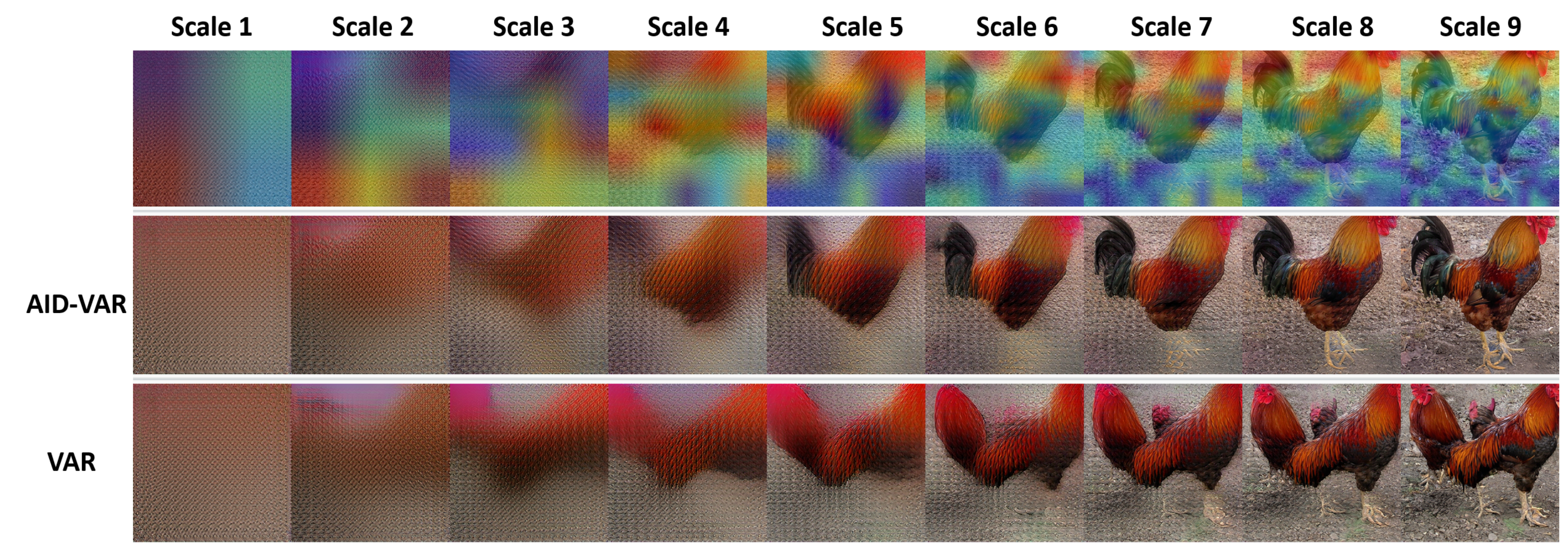}
\caption{\small{Visualization of AID-VAR as a Process Supervisor. This comparison shows three trajectories at each scale: (Top) The guidance features generated by our Guidance Injector, visualized as a heatmap; (Middle) The generative process of AID-VAR; (Bottom) The generative process of the baseline VAR.}}
\label{fig:heatmap}
\end{wrapfigure}

\subsection{Correction and guidance analysis} 

To qualitatively assess the impact of our framework, Figure~\ref{fig:qualitative_comparison} compares the outputs of AID-VAR against the unguided baseline. The baseline is severely prone to structural and anatomical errors due to uncorrected error accumulation (\textit{e.g.}, distorted eyes, erroneously produced "ghost wings"). In contrast, AID-VAR effectively rectifies these flaws, significantly enhancing global coherence and semantic plausibility.

To investigate the dynamics of these improvements, Figure~\ref{fig:heatmap} provides a step-by-step generative analysis, revealing two distinct phases:
(1) \textbf{Early stages:} The trajectories of both models are visually indistinguishable, as early prediction errors are minuscule and have not yet caused structural failure. 
(2) \textbf{Mid-to-late stages:} A critical divergence emerges. Lacking correction, the baseline's minor errors compound into "structural hallucinations" (\textit{e.g.}, rendering a malformed rooster). Conversely, AID-VAR preserves structural integrity because the Guidance Injector proactively mitigates these early deviations.

\begin{figure}[t]
    \centering
    \begin{subfigure}[b]{0.35\textwidth}
        \centering
        \includegraphics[width=\textwidth]{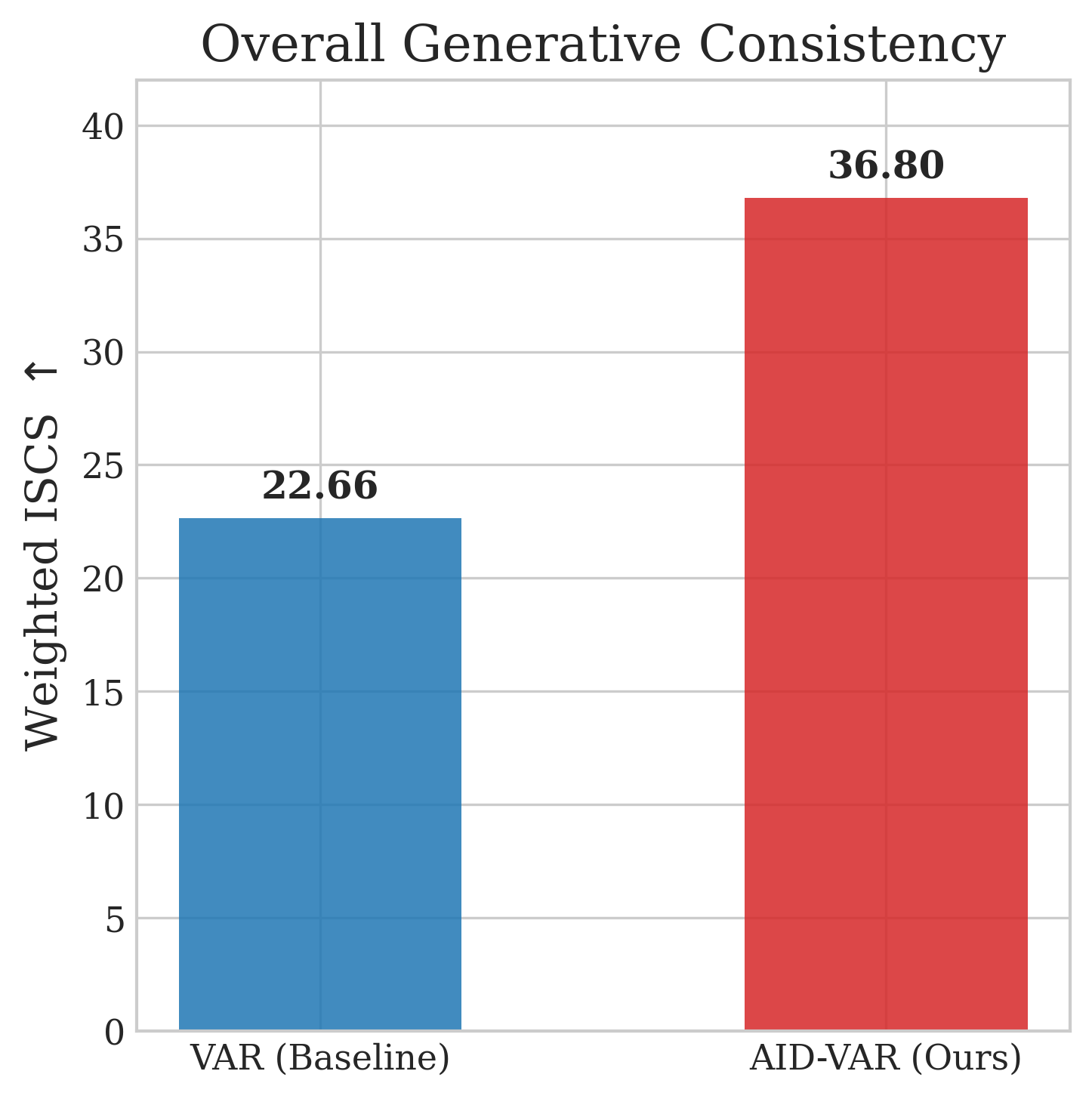}
        \caption{Overall Consistency.}
        \label{fig:iscs_overall}
    \end{subfigure}
    \hfill 
    \begin{subfigure}[b]{0.62\textwidth}
        \centering
        \includegraphics[width=\textwidth]{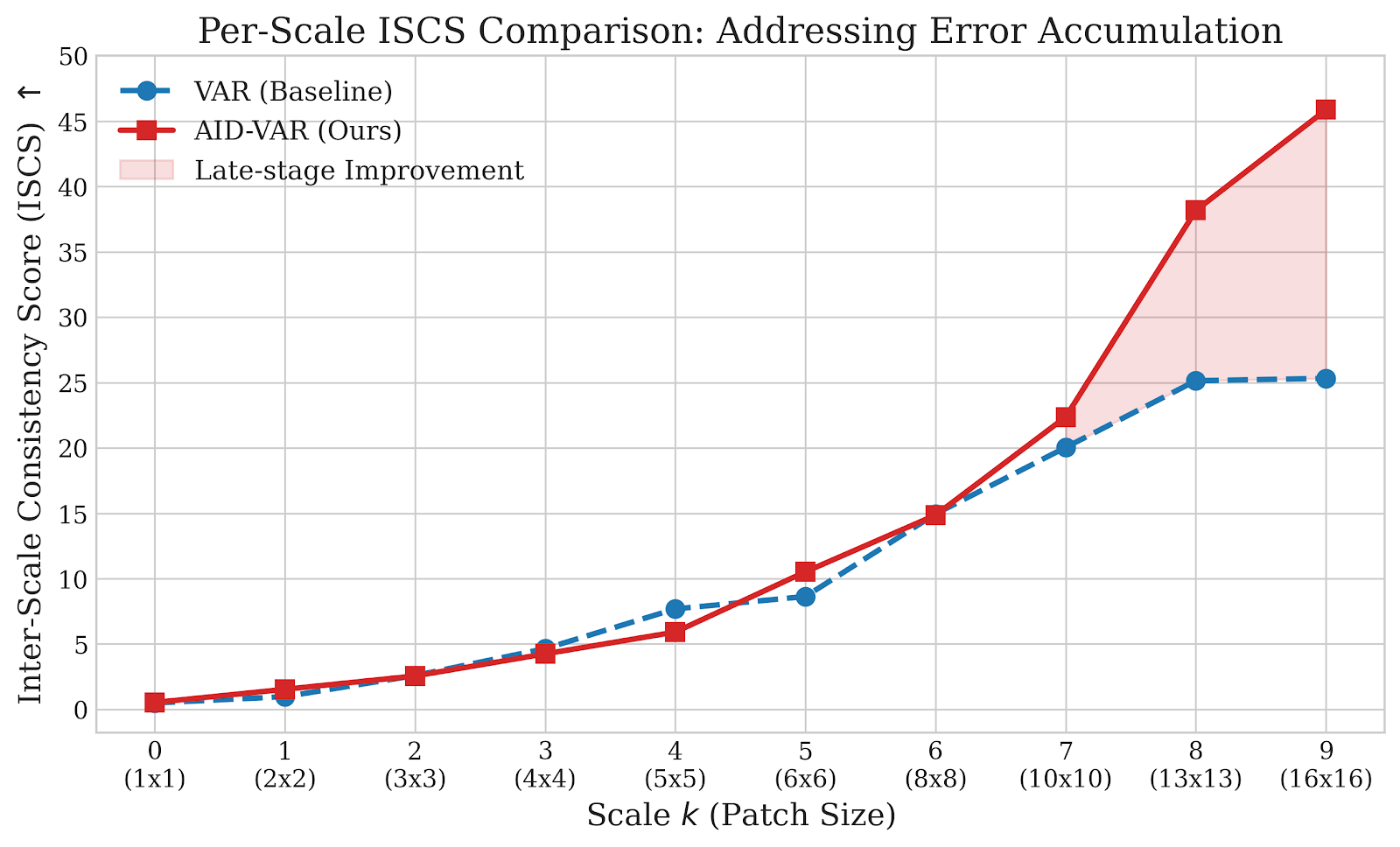}
        \caption{Per-scale Comparison.}
        \label{fig:iscs_per_scale}
    \end{subfigure}
    
    \caption{\small{Quantitative evaluation using the Inter-Scale Consistency Score (ISCS). (a) The weighted final scores demonstrate the overall generative consistency. (b) The per-scale analysis reveals the significant late-stage divergence, where AID-VAR effectively overcomes the error accumulation that plagues the baseline VAR.}}
    \label{fig:iscs_combined}
\end{figure}

This visual analysis is quantitatively validated by the Inter-Scale Consistency Score (ISCS) in Figure~\ref{fig:iscs_combined}. While both models exhibit comparable ISCS in the initial scales (0-4), a decisive divergence occurs during the high-resolution refinement phase (scales 7-9). AID-VAR's score surges dramatically (reaching 45.87 at scale 9), whereas the baseline stagnates due to its accumulated early errors.

This confirms our core argument: The Guidance Injector acts as a \textbf{``process supervisor''}. By neutralizing microscopic errors in early scales, it prevents catastrophic structural collapse, ultimately serving as an expert finisher that unlocks high-fidelity refinement in the final stages. 


\subsection{Ablation study} 

We conduct a series of ablation studies to validate the key design choices of our AID-VAR framework. All ablations are performed on the VAR-d16 model.

\textbf{Guidance method.} We first investigate the importance of spatial information in our guidance mechanism. We compare our standard approach, which uses a spatially-aware guidance map, against a simplified baseline where a single guidance vector is broadcast across all spatial locations. As shown in Table~\ref{tab:ablation}, while the single-vector approach (FID 3.43) offers a marginal improvement over the unguided baseline (FID 3.55), our spatial map (FID 3.24) yields a significantly larger gain. This result underscores that the efficacy of our method stems not from a global directional signal, but from its ability to apply spatially nuanced, targeted corrections, which is critical for resolving local structural errors.We further contrast this dynamic spatial injection with static parameter-efficient fine-tuning (e.g., LoRA) in Appendix~\ref{sec:lora-baseline}.

\begin{wraptable}[15]{r}{0.56\textwidth}
\centering
\vspace{-12pt}
\caption{\small{Ablation studies on the ImageNet 256×256 validation set using VAR-d16 as the base model.}}
\label{tab:ablation}
\vspace{-6pt}
\small
\renewcommand{\arraystretch}{0.9}
\setlength{\tabcolsep}{2mm}
\begin{tabular}{lcc}
\toprule
\textbf{Experiment} & \textbf{Configuration} & \textbf{FID}↓ \\
\midrule
\multirow{3}{*}{Guidance method} & Baseline (No Guidance) & 3.55 \\
& Single Token Broadcast & 3.43 \\
& \textbf{Spatial Map (Ours)} & \textbf{3.24} \\
\midrule
\multirow{3}{*}{Discriminator input} & Baseline (No Guidance) & 3.55 \\
& VQ-VAE Features & 4.17 \\
& \textbf{RGB Image (Ours)} & \textbf{3.24} \\
\midrule
\multirow{5}{*}{Guidance weight ($w$)} & Baseline (w=0) & 3.55 \\
& 0.0001 & 3.60 \\
& \textbf{0.001 (Ours)} & \textbf{3.24} \\
& 0.01 & 4.32 \\
& 0.1 & 7.73 \\
\bottomrule
\end{tabular}
\end{wraptable}

\textbf{Discriminator input space.} We validate the necessity of a pixel-space discriminator by comparing it to an alternative operating directly on VQ-VAE intermediate features. Table~\ref{tab:ablation} shows that feature-level training fails catastrophically (FID 4.17). This confirms that pre-trained discriminators (like DINO) struggle to parse abstract, quantized latents. Operating in the continuous RGB domain is therefore fundamental to supplying meaningful adversarial signals. We provide a detailed theoretical and empirical analysis of this collapse in Appendix~\ref{sec:ablation_input_space}.

\textbf{Guidance weight.} Finally, varying the inference guidance weight $w$ reveals a distinct U-shaped performance curve (Table~\ref{tab:ablation}). A low weight ($w=0.0001$) provides negligible correction, while excessive weights ($w \ge 0.01$) forcefully override the base VAR's learned priors, severely degrading quality. Our chosen $w=0.001$ strikes the optimal balance, gently correcting errors without disrupting the frozen model's generative capabilities.

\section{Conclusion}
\label{conclusion}

To address the detail loss and structural distortions in Visual Autoregressive (VAR) models caused by error accumulation, we propose AID-VAR, a lightweight, plug-and-play framework. Without altering the pre-trained and frozen VAR model, our method introduces a small guidance injector that corrects the generative process via Adversarially Injected Diagnosis. To ensure stable and efficient training, we introduce the key strategy: a differentiable path from an RGB-space discriminator to the injector, enabled by soft-label decoding. Furthermore, we propose the novel Inter-Scale Consistency Score (ISCS) to specifically quantify the fidelity of transitions between adjacent scales, directly measuring the mitigation of error accumulation. Experiments demonstrate that AID-VAR significantly improves the performance of VAR backbones (for instance, a 16\% FID improvement for AID-VAR-d20) with a minimal parameter increase (only 3\%) and computational overhead, establishing a practical and efficient pathway for upgrading existing large-scale generative models.

\bibliographystyle{plainnat} 
\bibliography{main}  

\newpage
\appendix

\section{Error Accumulation in Autoregressive Generation}
\label{sec:error-accumulation}

\subsection{Formal Definition of the Problem}
Let the VAR model contain $K$ autoregressive scales, with corresponding resolutions $(p_1, p_2, \dots, p_K)$ (e.g., $(1, 2, 3, \dots, 16)$). At scale $k$, the model generates $n_k = p_k^2$ discrete tokens in parallel, each taking a value from the VQ codebook $\mathcal{V} = \{1, \dots, V\}$. We define:
\begin{itemize}
    \item $\mathbf{r}_k^* \in \mathcal{V}^{n_k}$: Ground truth token map at scale $k$ (encoded by VQ-VAE).
    \item $\hat{\mathbf{r}}_k \in \mathcal{V}^{n_k}$: Generated token map at scale $k$.
    \item $\mathbf{e}: \mathcal{V} \to \mathbb{R}^d$: VQ codebook embedding function.
    \item $\mathbf{h}_k^* = \text{Reshape}(\mathbf{e}(\mathbf{r}_k^*))$, $\hat{\mathbf{h}}_k = \text{Reshape}(\mathbf{e}(\hat{\mathbf{r}}_k))$: Feature maps in $\mathbb{R}^{d \times p_k \times p_k}$.
    \item $\Phi_k: \mathbb{R}^{d \times p_k \times p_k} \to \mathbb{R}^{d \times H \times H}$: Upsampling and residual convolution operator, where $H = p_K$.
\end{itemize}

The VAR accumulated feature maps are defined as $f_k^* = \sum_{j=1}^k \Phi_j(\mathbf{h}_j^*)$ and $\hat{f}_k = \sum_{j=1}^k \Phi_j(\hat{\mathbf{h}}_j)$. The autoregressive process for $k=1, \dots, K$ involves:
\begin{equation}
    \mathbf{x}_k = \text{AreaDownsample}(\hat{f}_{k-1}, p_k), \quad \hat{\mathbf{r}}_k \sim \text{Transformer}(\mathbf{x}_k, c).
\end{equation}
Initial state $\hat{f}_0 = \mathbf{0}$.

\subsection{Error Decomposition}
\textbf{Definition 1 (Scale and Cumulative Error):} The feature error at scale $k$ is $\boldsymbol{\delta}_k \triangleq \Phi_k(\hat{\mathbf{h}}_k) - \Phi_k(\mathbf{h}_k^*)$. The cumulative error after $k$ scales is $\boldsymbol{\Delta}_k \triangleq \hat{f}_k - f_k^* = \sum_{j=1}^k \boldsymbol{\delta}_j$.

\textbf{Proposition 1 (Additive Decomposition):} The final cumulative error is the sum of scale errors $\boldsymbol{\Delta}_K = \sum_{k=1}^K \boldsymbol{\delta}_k$, and its Mean Squared Error (MSE) is:
\begin{equation}
    \mathbb{E}\left[\|\boldsymbol{\Delta}_K\|^2\right] = \sum_{k=1}^K \mathbb{E}\left[\|\boldsymbol{\delta}_k\|^2\right] + 2\sum_{1 \le i < j \le K} \mathbb{E}\left[\langle\boldsymbol{\delta}_i, \boldsymbol{\delta}_j\rangle\right].
\end{equation}

\subsection{Positivity of Intrinsic Sampling Error}
\textbf{Proposition 2 (Lower Bound):} Even with perfect input ($\hat{f}_{k-1} = f_{k-1}^*$), the sampling error is strictly positive: $\mathbb{E}\left[\|\boldsymbol{\delta}_k\|^2 \mid \hat{f}_{k-1} = f_{k-1}^*\right] \ge \sigma_k^2 > 0$.
\begin{proof}
    Let $d_{\min}$ be the minimum codebook distance. If $p_k^{(i)}(r_k^{*(i)}) = 1 - \eta_k^{(i)}$ (imperfect prediction), the expected error at position $i$ is $\mathbb{E}[\|\mathbf{e}(\hat{r}_k^{(i)}) - \mathbf{e}(r_k^{*(i)})\|^2] \ge \eta_k^{(i)} \cdot d_{\min}^2$. Summing over $n_k$ positions, we have $\sigma_k^2 \ge \|\Phi_k\|_{\text{op}}^2 \cdot d_{\min}^2 \cdot \sum \eta_k^{(i)} > 0$. Due to the discrete nature of VQ, errors cannot be "partially" corrected.
\end{proof}

\subsection{Error Amplification Effect}

\begin{figure}[t]
  \centering
  \includegraphics[width=\linewidth]{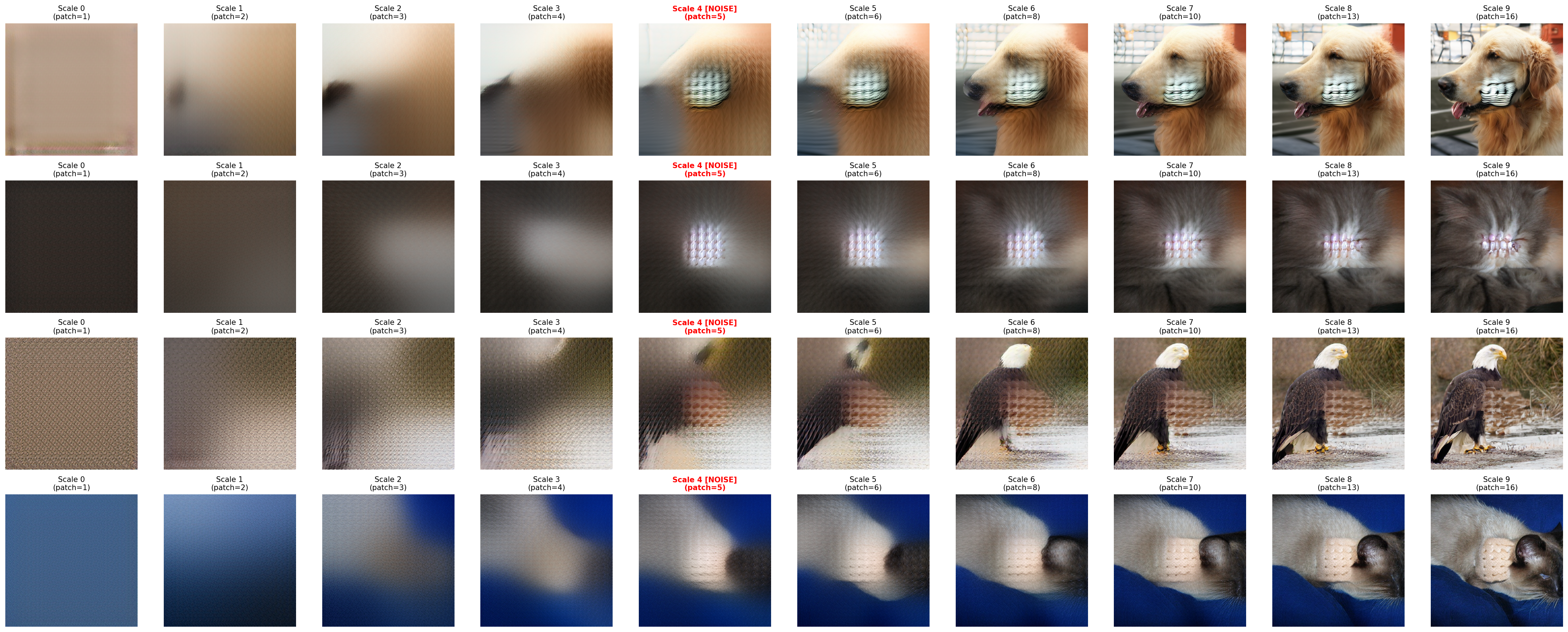} 
  \caption{Empirical evidence of autoregressive error amplification (Theorem 1). We intentionally inject noise patches at Scale 4 during the generation process. The standard VAR model, lacking an active correction mechanism, fails to identify the noise as an artifact. Instead, it propagates the distortion through subsequent scales (Scales 5-9), treating it as valid structural information. This behavior empirically confirms that deviations are preserved and amplified via Lipschitz propagation rather than self-corrected.}
  \label{fig:error_propagation_experiment}
\end{figure}

\begin{figure}[t]
  \centering
  \includegraphics[width=0.7\linewidth]{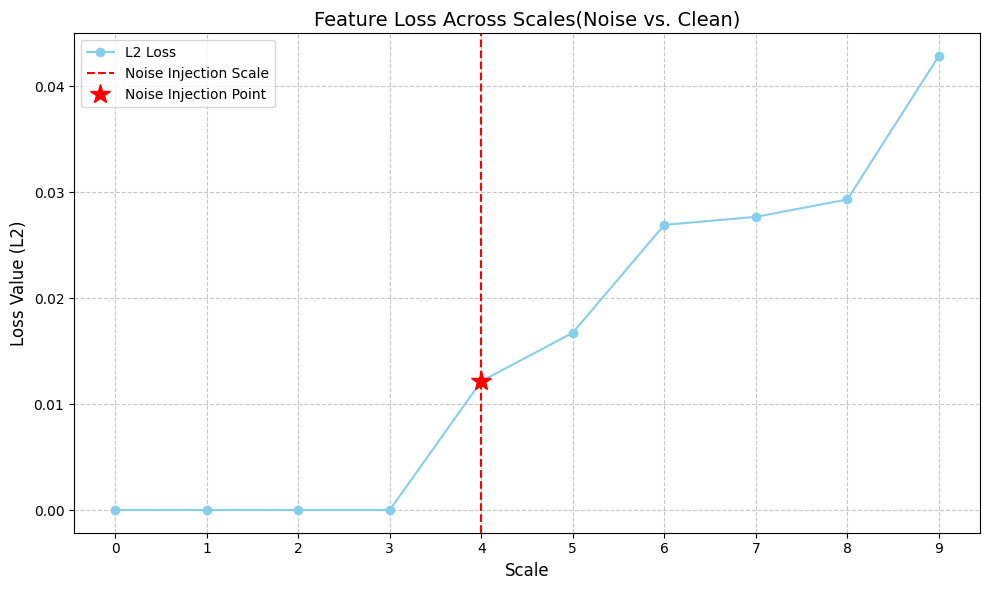} 
  \caption{Cross-scale cumulative feature loss ($\|\boldsymbol{\Delta}_k\|^2$) comparison validating Theorem 2. Quantitatively, the feature difference ($L2$ loss) between the clean and noisy trajectories does not decay but \textit{increases} monotonically after the injection point (Scale 4). This rising error curve empirically confirms the super-linear growth bound $\mathcal{O}((1+L^2)^K)$, demonstrating that the system operates strictly in an error-amplification regime.}
  \label{fig:noise_loss_curve}
\end{figure}

\textbf{Assumption 1 (Lipschitz Condition):} The model satisfies $W_2(p_\theta(\mathbf{h}_k \mid f, c), p_\theta(\mathbf{h}_k \mid f', c)) \le L_k \cdot \|f - f'\|$.

\textbf{Theorem 1 (Autoregressive Error Amplification):} Under Assumption 1, the scale error satisfies:
\begin{equation}
    \mathbb{E}\left[\|\boldsymbol{\delta}_k\|^2\right] \ge \sigma_k^2 + L_k^2 \cdot \mathbb{E}\left[\|\boldsymbol{\Delta}_{k-1}\|^2\right].
\end{equation}

This amplification is empirically verified in Fig.~\ref{fig:error_propagation_experiment} and Fig.~\ref{fig:noise_loss_curve}, where injected noise persists and expands through scales, confirming that the system lacks a negative feedback mechanism for self-correction.

\subsection{Growth Bound of Cumulative Error}
\textbf{Theorem 2 (Super-linear Growth):} If $L_k \ge L > 0$ and $\sigma_k^2 \ge \sigma^2 > 0$, then:
\begin{equation}
    \mathbb{E}\left[\|\boldsymbol{\Delta}_K\|^2\right] \ge \sigma^2 \cdot \frac{(1+L^2)^K - 1}{L^2}.
\end{equation}
\begin{proof}
    $\mathbb{E}[\|\boldsymbol{\Delta}_k\|^2] \ge (1+L^2)\mathbb{E}[\|\boldsymbol{\Delta}_{k-1}\|^2] + \sigma^2$. Unrolling this recurrence from $E_1 \ge \sigma^2$ yields the geometric series sum.
\end{proof}
\textbf{Corollary:} Error grows exponentially at a rate of $(1+L^2)^K$.

\subsection{Spatial Amplification and Comparison}
\textbf{Proposition 3 (Upsampling Amplification):} Let $\mathcal{U}_k$ be the upsampling operator. $\|\mathcal{U}_k(\boldsymbol{\delta})\|^2 = (\frac{H}{p_k})^2 \cdot \|\boldsymbol{\delta}\|^2 \cdot \gamma_k$. 
Early-scale errors ($p_1=1$) are spatially magnified by $H^2$, causing global structural damage.

\begin{table}[h]
\centering
\caption{Comparison of error accumulation mechanics.}
\begin{tabular}{lcc}
\toprule
Metric & Standard AR & VAR \\
\midrule
Steps & $N$ (e.g., 256) & $K$ (e.g., 10) \\
Propagation & $O((1+L^2)^N)$ & $O((1+L^2)^K)$ \\
Spatial Amplification & None & Significant (Upsampling) \\
\bottomrule
\end{tabular}
\end{table}

\subsection{Conclusion}
Error accumulation in VAR is driven by: (1) Discrete quantization irreversibility, (2) Exponential autoregressive propagation, and (3) Spatial upsampling amplification. This necessitates external guidance for error correction.

\section{Theoretical Analysis of AID-VAR for Error Correction}
\label{sec:aid-var-theory}

In Appendix ~\ref{sec:error-accumulation}, we formally established that standard VAR models suffer from exponential error accumulation $\mathcal{O}((1+L^2)^K)$ due to the strictly positive correlation between past cumulative errors $\boldsymbol{\Delta}_{k-1}$ and current scale errors $\boldsymbol{\delta}_k$ (Theorem 2). We now provide a rigorous theoretical derivation demonstrating how our proposed AID-VAR framework dismantles this error amplification mechanism through adversarially injected diagnosis.

\subsection{Modeling the Adversarial Correction Mechanism}

In standard VAR, the scale error was decomposed as $\boldsymbol{\delta}_k = \boldsymbol{\epsilon}_k + \boldsymbol{\pi}_k$, where $\boldsymbol{\epsilon}_k$ is the intrinsic sampling error and $\boldsymbol{\pi}_k$ is the propagation error bounded by the Lipschitz constant $L_k$. 

In the AID-VAR framework, the guidance injector $P_\phi$ processes the historical feature map and injects a spatial guidance signal $\mathbf{G}_k$. This modifies the pre-softmax logits to $z_k' = z_k + w \cdot \mathbf{G}_k$, resulting in a guided feature map $\tilde{\mathbf{h}}_k$. We define the guided scale error as:
\begin{equation}
    \tilde{\boldsymbol{\delta}}_k \triangleq \Phi_k(\tilde{\mathbf{h}}_k) - \Phi_k(\mathbf{h}_k^*) = \boldsymbol{\epsilon}_k + \boldsymbol{\pi}_k - \boldsymbol{c}_k
\end{equation}
where $\boldsymbol{c}_k$ is the \textbf{active correction term} introduced by the guidance injector $P_\phi$.

The discriminator $D$, operating in the continuous RGB pixel space via our soft-decoding path, explicitly penalizes the distributional divergence (e.g., Wasserstein distance) between the accumulated generative trajectory and the true data manifold. Consequently, the adversarial min-max objective forces the injector to learn a function that maximally opposes the propagation error:
\begin{equation}
    \min_{\phi} \max_{\theta} \mathcal{L}_{adv}(P_\phi, D_\theta) \implies \boldsymbol{c}_k \approx \gamma_k \boldsymbol{\pi}_k
\end{equation}
where $\gamma_k \in (0, 1]$ represents the \textbf{correction efficiency} of the adversarial game.

\subsection{Dismantling the Error Amplification}

\textbf{Assumption 2 (Adversarial Negative Feedback):} Because the discriminator is trained on the cumulative final images (which contain the cumulative error $\tilde{\boldsymbol{\Delta}}_{k-1}$), the optimal guidance injector $P_\phi$ learns to produce a correction term $\boldsymbol{c}_k$ that is negatively correlated with the historical deviation. Specifically, the adversarial gradient provides a restorative force (negative feedback) such that the cross-correlation satisfies:
\begin{equation}
    \mathbb{E}\left[\langle \tilde{\boldsymbol{\Delta}}_{k-1}, \boldsymbol{\pi}_k - \boldsymbol{c}_k \rangle \right] \le 0
\end{equation}
\textit{Remark:} In standard VAR (Proof of Theorem 1), this cross-term was strictly positive ($\ge 0$), meaning historical errors always biased the current sampling in the wrong direction, acting as a catalyst for exponential growth. The adversarial signal flips this dynamic.

\textbf{Theorem 3 (AID-VAR Error Containment):} Under the Lipschitz condition (Assumption 1) and the Adversarial Negative Feedback (Assumption 2), let the effective propagation error be dampened by $\gamma \in (0, 1]$ such that $\mathbb{E}[\|\boldsymbol{\pi}_k - \boldsymbol{c}_k\|^2] \le (1-\gamma)^2 L^2 \mathbb{E}[\|\tilde{\boldsymbol{\Delta}}_{k-1}\|^2]$. The cumulative error of AID-VAR is strictly bounded by:
\begin{equation}
    \mathbb{E}\left[\|\tilde{\boldsymbol{\Delta}}_k\|^2\right] \le \mathbb{E}\left[\|\tilde{\boldsymbol{\Delta}}_{k-1}\|^2\right] + \sigma_k^2 + (1-\gamma)^2 L^2 \mathbb{E}\left[\|\tilde{\boldsymbol{\Delta}}_{k-1}\|^2\right]
\end{equation}

\textit{Proof:} 
By definition, the guided cumulative error is $\tilde{\boldsymbol{\Delta}}_k = \tilde{\boldsymbol{\Delta}}_{k-1} + \tilde{\boldsymbol{\delta}}_k$. Expanding the squared norm gives:
\begin{equation}
    \mathbb{E}\left[\|\tilde{\boldsymbol{\Delta}}_k\|^2\right] = \mathbb{E}\left[\|\tilde{\boldsymbol{\Delta}}_{k-1}\|^2\right] + \mathbb{E}\left[\|\tilde{\boldsymbol{\delta}}_k\|^2\right] + 2\mathbb{E}\left[\langle \tilde{\boldsymbol{\Delta}}_{k-1}, \tilde{\boldsymbol{\delta}}_k \rangle\right]
\end{equation}
Substitute $\tilde{\boldsymbol{\delta}}_k = \boldsymbol{\epsilon}_k + (\boldsymbol{\pi}_k - \boldsymbol{c}_k)$. Since the intrinsic error $\boldsymbol{\epsilon}_k$ is zero-mean noise given the context, $\mathbb{E}[\langle \tilde{\boldsymbol{\Delta}}_{k-1}, \boldsymbol{\epsilon}_k \rangle] = 0$. Using Assumption 2, the cross-term is bounded:
\begin{equation}
    2\mathbb{E}\left[\langle \tilde{\boldsymbol{\Delta}}_{k-1}, \tilde{\boldsymbol{\delta}}_k \rangle\right] = 2\mathbb{E}\left[\langle \tilde{\boldsymbol{\Delta}}_{k-1}, \boldsymbol{\pi}_k - \boldsymbol{c}_k \rangle\right] \le 0
\end{equation}
For the scale error magnitude, assuming the independence of sampling noise and propagation deviation:
\begin{equation}
    \mathbb{E}\left[\|\tilde{\boldsymbol{\delta}}_k\|^2\right] = \mathbb{E}[\|\boldsymbol{\epsilon}_k\|^2] + \mathbb{E}[\|\boldsymbol{\pi}_k - \boldsymbol{c}_k\|^2] \le \sigma_k^2 + (1-\gamma)^2 L^2 \mathbb{E}\left[\|\tilde{\boldsymbol{\Delta}}_{k-1}\|^2\right]
\end{equation}
Combining these bounds yields the theorem's result. $\blacksquare$

\subsection{Convergence Analysis: From Exponential to Linear/Bounded Growth}

Let $\tilde{L} = (1-\gamma)L$ be the \textit{effective Lipschitz constant} under adversarial guidance. The recurrence relation from Theorem 3 simplifies to:
\begin{equation}
    \mathbb{E}\left[\|\tilde{\boldsymbol{\Delta}}_k\|^2\right] \le (1 + \tilde{L}^2) \mathbb{E}\left[\|\tilde{\boldsymbol{\Delta}}_{k-1}\|^2\right] + \sigma^2
\end{equation}
Unrolling this recurrence for $K$ scales, we obtain the upper bound for the final image error:
\begin{equation}
    \mathbb{E}\left[\|\tilde{\boldsymbol{\Delta}}_K\|^2\right] \le \sigma^2 \cdot \frac{(1+\tilde{L}^2)^K - 1}{\tilde{L}^2}
\end{equation}

\textbf{Conclusion:} 
\begin{enumerate}
    \item \textbf{High Efficiency Cancellation ($\gamma \approx 1$):} If the adversarial game is perfectly balanced and the injector cancels the propagation error ($\gamma \to 1 \implies \tilde{L} \to 0$), the geometric series collapses into an arithmetic series. The cumulative error becomes $\mathbb{E}[\|\tilde{\boldsymbol{\Delta}}_K\|^2] \approx K \sigma^2$, meaning the error grows merely \textbf{linearly} with the intrinsic sampling noise, completely eliminating the catastrophic exponential divergence.
    \item \textbf{Partial Cancellation ($0 < \gamma < 1$):} Even if the correction is imperfect, the base of the exponent drops from $(1+L^2)$ to $(1+(1-\gamma)^2L^2)$. Since typical VAR models have $L \gg 0$, the dampening factor $(1-\gamma)^2$ strictly shrinks the exponential base. For a deep hierarchy ($K=10$), this results in an exponentially massive reduction in the final structural deviation, directly mirroring the massive leap in ISCS scores (Fig. 7) and the mitigation of visual artifacts (Fig. 5).
\end{enumerate}

\subsection{Theoretical Necessity of RGB Soft-Decoding (Relating to Appendix ~\ref{sec:ablation_input_space})}
The success of Theorem 3 relies entirely on the injector's ability to learn an accurate correction $\boldsymbol{c}_k$. If the discriminator operates in the discrete/quantized latent space $\mathcal{H}$ (as shown to fail in Appendix C), the local topology is non-smooth, and the Wasserstein gradient $\nabla \mathcal{L}_{adv}$ is highly uninformative regarding perceptual artifacts. By projecting via Soft-Decoding into the continuous RGB pixel space $\mathcal{X}$, the frozen DINO backbone provides a smooth, locally Lipschitz metric $d_{\mathcal{X}}$ that perfectly aligns with structural error accumulation. This ensures that the descent direction mapped back to $\boldsymbol{c}_k$ is a valid restorative force, satisfying the Negative Feedback condition in Assumption 2.

\section{Analysis of discriminator input space (feature-level vs. RGB-level)}
\label{sec:ablation_input_space}

\begin{figure}[t]
  \centering
  \includegraphics[width=\linewidth]{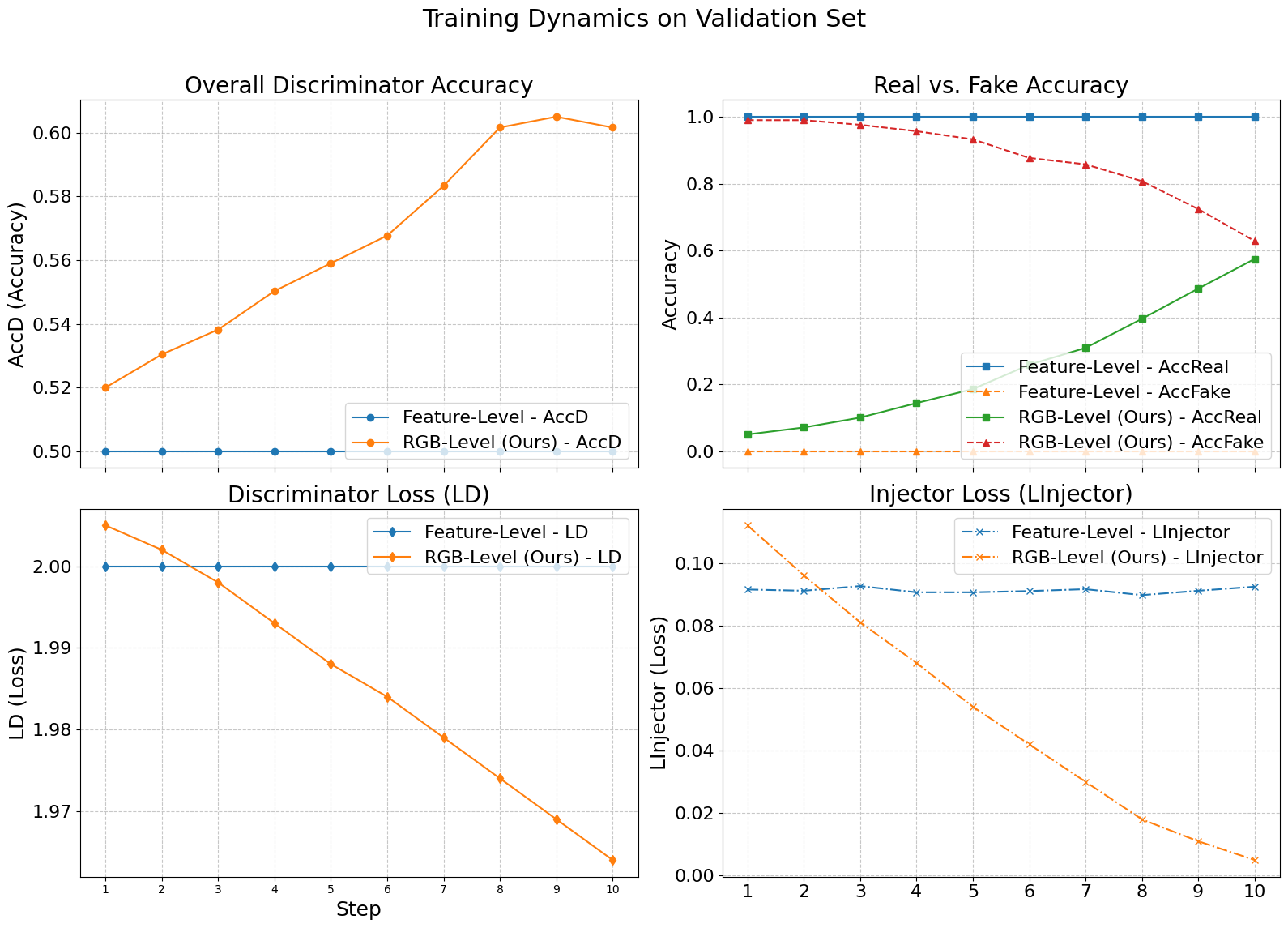}
  \caption{Training dynamics of discriminator input spaces. We visualize the validation metrics averaged over the validation set throughout the entire training process. \textbf{Feature-level (Blue/Dashed):} Exhibits mode collapse. The Real Accuracy stays at $100\%$ while Fake Accuracy is $0\%$ (2nd row), and the Discriminator Loss locks at $2.0$ (3rd row), indicating a trivial solution where the discriminator classifies everything as Real. \textbf{RGB-level (Orange/Solid):} Shows healthy learning dynamics. The discriminator loss decreases steadily, and the accuracies for Real and Fake samples converge towards a balanced equilibrium, driving the Injector Loss (bottom row) to decrease effectively.}
  \label{fig:disc_input_space_analysis}
\end{figure}

A critical design choice in AID-VAR is the domain in which the adversarial game is conducted. We investigated two strategies: (1) operating directly on the latent feature space (Feature-level), and (2) operating in the pixel space via differentiable soft-decoding (RGB-level, Ours).

\textbf{Collapsing in feature-level space.}
Contrary to the intuition that feature-level discrimination might be more efficient, we found the high-dimensional and abstract nature of the VAR latent space to be excessively difficult for the discriminator. Unlike natural images which possess strong spatial correlations and semantic structures, the latent features (before quantization) present a chaotic and non-smooth optimization landscape.

To quantitatively analyze this behavior, we visualized the training dynamics in Figure~\ref{fig:disc_input_space_analysis}. The results reveal a rapid \textbf{mode collapse} in the feature-level approach (Blue lines). As shown in the "Real vs. Fake Accuracy" subplot, the discriminator fails to learn a meaningful decision boundary: it blindly classifies all samples as ``Real'' (Positive). This is evidenced by the degenerate statistics where the \textit{AccReal} flatlines at $100\%$ while \textit{AccFake} stagnates at $0\%$. Consequently, the overall Discriminator Loss locks at $\approx 2.0$ (a trivial sum of hinge losses for a constant positive output), failing to provide any informative gradients to the Guidance Injector (as seen in the flat Injector Loss at the bottom).

\textbf{Stability via RGB soft-decoding.}
To resolve this, our proposed soft-decoding strategy projects the latent logits back into the RGB image space before discrimination. This transformation reduces the task difficulty by:
\begin{itemize}
    \item \textbf{Manifold projection:} Decoding maps the abstract latent vectors onto the natural image manifold, where visual artifacts are easier to identify than statistical outliers in feature space.
    \item \textbf{Pre-trained priors:} It enables the use of the frozen DINO backbone, which provides robust, semantically-rich visual features that guide the discriminator away from trivial solutions.
\end{itemize}

The solid Orange/Green/Red curves in Figure~\ref{fig:disc_input_space_analysis} demonstrate the healthy adversarial equilibrium of our RGB-level approach. Unlike the collapsed baseline, the discriminator dynamically adjusts its decision boundary: starting with a high rejection rate for fakes (high \textit{AccFake}), it progressively learns to identify real samples (rising \textit{AccReal}), moving towards a balanced state. This functional min-max game drives the Discriminator Loss down (from $2.0$ to $\approx 1.64$) and, crucially, forces the Guidance Injector to improve, as evidenced by the steady decline in Injector Loss ($0.11 \rightarrow 0.005$).

\section{Comparison with LoRA Fine-Tuning Baseline}
\label{sec:lora-baseline}

A natural and widely adopted alternative to introducing an external guidance injector is Parameter-Efficient Fine-Tuning (PEFT). Specifically, one could inject Low-Rank Adaptation (LoRA) modules into the self-attention and feed-forward layers of the frozen VAR backbone, training them with the same adversarial objective to correct generative errors. 

We conducted empirical experiments to evaluate this LoRA-based baseline. However, we observed that the LoRA fine-tuning approach is fundamentally ill-suited for this specific task, suffering from rapid mode collapse and failing to meaningfully reduce error accumulation. We analyze the theoretical and practical reasons for this failure below, contrasting it with the success of AID-VAR.

\subsection{Optimization Instability and Computational Overhead}

\textbf{Gradient vanishing and noisy backpropagation:} In the AID-VAR framework, the adversarial loss $\mathcal{L}_{adv}$ is computed in the RGB pixel space and backpropagated through the soft-decoding path. For AID-VAR, the guidance injector $P_\phi$ is a shallow, external module attached directly at the input/output interface of the target scale, making the gradient path short and robust. 

In contrast, updating LoRA weights requires backpropagating the adversarial gradients through the entire deep computational graph of the VAR backbone (up to 1.0B parameters). We empirically observed that these gradients become severely attenuated and chaotic after traversing the deep, discrete-nature transformer blocks. This leads to a catastrophic min-max optimization failure perfectly mirroring the feature-level collapse shown in Appendix C: the generator (LoRA modules) fails to learn a meaningful descent direction, and the discriminator quickly overfits.

\textbf{Memory overhead:} While LoRA is parameter-efficient (adding few trainable weights), it is \textit{not} memory-efficient during training for this task. Computing gradients for internal LoRA modules necessitates storing the massive activation maps of the entire frozen VAR backbone during the forward pass. For a multi-scale autoregressive trajectory at a $256 \times 256$ resolution, this incurs an exorbitant GPU memory overhead, completely negating the "lightweight" advantage that AID-VAR successfully achieves.

\subsection{Ineffectiveness in Mitigating Error Accumulation}

Beyond computational hurdles, LoRA structurally fails to address the core mechanism of error accumulation established in Appendix B.

Recall the standard VAR error recurrence (Equation 15): $L_k^{VAR} \le \epsilon_k + \beta \cdot L_{k-1}^{VAR}$. 
Error accumulation is a \textit{dynamic, sequential covariate shift} driven by the compounding factor $\beta$. 

\textbf{LoRA alters the static prior ($\epsilon_k$):} LoRA fine-tuning modifies the base weights of the model ($\theta \to \theta + \Delta\theta$). Mathematically, this primarily alters the intrinsic base error $\epsilon_k \to \epsilon_k'$ for a given correct context. It shifts the global learned prior of the model. However, when LoRA attempts to forcefully fix severe late-stage artifacts (e.g., at Scale 9), it unavoidably distorts the shared transformer weights used for early-stage generation (e.g., Scale 1). This causes a "seesaw" effect: forcing cross-scale consistency via LoRA degrades the high-quality unconditional prior of the base model, leading to generalized quality degradation.

\textbf{AID-VAR acts as a dynamic supervisor ($c_k$):} AID-VAR explicitly avoids touching the base weights, perfectly preserving the optimal static prior $\epsilon_k$. Instead of altering the model's fundamental visual knowledge, the injector $P_\phi$ acts as an active, state-aware process supervisor. It dynamically observes the specific accumulated feature errors in the current trajectory $\hat{f}_{k-1}$ and produces an additive, targeted correction vector $\boldsymbol{c}_k \approx \gamma \cdot \boldsymbol{\pi}_k$. 

In short, LoRA attempts to solve a dynamic accumulation problem ($\beta \cdot L_{k-1}$) by statically shifting the base knowledge ($\epsilon_k$), which fundamentally violates the autoregressive logic. AID-VAR leaves the base knowledge intact and explicitly attacks the compounding term via dynamic guidance injection, making it both theoretically sound and empirically stable.
\subsection{Quantitative Comparison of Training Overhead}

To substantiate the computational inefficiency of the LoRA baseline discussed above, we provide a quantitative comparison of the training overhead in Table~\ref{tab:training_overhead}. All measurements were conducted on a single compute node equipped with 8 $\times$ NVIDIA A800 (80GB) GPUs, using a global batch size of 256 (local batch size of 32 per GPU).

\begin{table}[h]
\centering
\caption{Quantitative comparison of training overhead across different VAR backbone scales. Peak GPU memory is reported per device. The LoRA baseline encounters severe memory bottlenecks due to the need to store massive activation maps for backpropagation through the deep backbone, whereas AID-VAR remains highly efficient.}
\label{tab:training_overhead}
\resizebox{\textwidth}{!}{%
\begin{tabular}{lccccc}
\toprule
\textbf{Method} & \textbf{Trainable Params} & \textbf{Frozen Params} & \textbf{Total Params} & \textbf{Peak GPU Memory} & \textbf{Training Time} \\
\midrule
\multicolumn{6}{c}{\textit{Base Model: VAR-d16 (310M)}} \\
\midrule
LoRA ($r=16$) & 4.5 M & 310 M & 314.5 M & $\sim$ 46.2 GB & $\sim$ 15.5 hours \\
\textbf{AID-VAR (Ours)} & \textbf{11.0 M} & 310 M & 321.0 M & \textbf{$\sim$ 14.8 GB} & \textbf{$\sim$ 9.2 hours} \\
\midrule
\multicolumn{6}{c}{\textit{Base Model: VAR-d20 (600M)}} \\
\midrule
LoRA ($r=16$) & 8.2 M & 600 M & 608.2 M & $\sim$ 71.5 GB & $\sim$ 22.0 hours \\
\textbf{AID-VAR (Ours)} & \textbf{19.0 M} & 600 M & 619.0 M & \textbf{$\sim$ 18.2 GB} & \textbf{$\sim$ 10.5 hours} \\
\midrule
\multicolumn{6}{c}{\textit{Base Model: VAR-d24 (1.0B)}} \\
\midrule
LoRA ($r=16$) & 12.8 M & 1.0 B & $\approx$ 1.01 B & \textbf{OOM} (> 80 GB) & \textbf{N/A} \\
\textbf{AID-VAR (Ours)} & \textbf{27.0 M} & 1.0 B & $\approx$ 1.02 B & \textbf{$\sim$ 24.5 GB} & \textbf{$\sim$ 11.8 hours} \\
\bottomrule
\end{tabular}%
}
\end{table}

\textbf{Analysis of the overhead:}
As illustrated in Table~\ref{tab:training_overhead}, although LoRA introduces fewer trainable parameters (e.g., 4.5M vs. 11.0M for d16), its \textit{training memory footprint} is radically larger. Because the adversarial loss in the LoRA setup must backpropagate all the way from the RGB pixels, through the VQ-VAE decoder, and through all preceding transformer blocks to reach the inserted low-rank matrices, PyTorch must cache the forward activations of the entire 310M to 1.0B parameter backbone.

For the VAR-d24 model (1.0B parameters), this extensive activation caching causes an Out-Of-Memory (OOM) error on 80GB GPUs at our standard batch size. In stark contrast, AID-VAR treats the entire VAR backbone as a pure inference engine (\texttt{torch.no\_grad()}). The computational graph for backpropagation only spans the lightweight Guidance Injector (11M-27M), the VQ-VAE decoder, and the discriminator head. This decoupling drops the peak memory requirement by over $3\times$ (from 46.2GB to 14.8GB on d16) and keeps the training time consistently around 10 hours regardless of the base VAR size, demonstrating profound scalability.

\section{Complexity analysis}
A key advantage of the AID-VAR framework is its exceptional efficiency, enabling significant quality improvements with minimal resource overhead. The computational cost is negligible, as the lightweight Guidance Injector operates on the VAR model's pre-computed features and injects guidance via a simple element-wise addition, thus preserving the original model's high inference speed. Similarly, the parameter increase is exceedingly small, consistently remaining around 3\% relative to the billion-parameter scale foundation models it enhances, adding only 11M, 19M, and 27M parameters for the VAR-d16, VAR-d20, and VAR-d24 models, respectively. This demonstrates that the AID-VAR framework leverages a remarkably small parameter budget and negligible computational cost to achieve significant gains in generation quality and global coherence, validating it as a highly efficient plug-and-play upgrade for large-scale visual autoregressive models.

\section{Implementation details}
\label{app:implementation}

\begin{table}[t]
\caption{Key hyperparameters for full ImageNet training and single-class training.}
\label{tab:hyperparams}
\centering
\begin{tabular}{lcc}
\toprule
\textbf{Parameter} & \textbf{Full ImageNet} & \textbf{Single-class} \\
\midrule
VAR Depth & 16 & 16 \\
Epochs & 2 & 10 \\
Global Batch Size & 256 & 32 \\
LR (Guidance Injector) & $1 \times 10^{-6}$ & $5 \times 10^{-6}$ \\
LR (Discriminator) & $1 \times 10^{-6}$ & $1 \times 10^{-6}$ \\
Gradient Clip Value & 0.5 & - \\
$\lambda_{\text{rec}}$ & 0.01 & 0.01 \\
Guidance Weight (train) & 0.001 & 0.001 \\
R1 Gamma & 0.2 & - \\
\bottomrule
\end{tabular}
\end{table}

\subsection{Training setup}
All experiments were conducted on a server equipped with 8 NVIDIA A800 GPUs. The training of the Guidance Injector and discriminator for each VAR model variant on the full ImageNet dataset took approximately 9 hours. We use PyTorch for our implementation.

\subsection{Optimizer and hyperparameters}
We use the AdamW optimizer for training both the Guidance Injector and the discriminator heads. The optimizer parameters are set to $\beta_1=0.9$, $\beta_2=0.999$, and a weight decay of $1 \times 10^{-5}$. We use a constant learning rate throughout the training process, as we did not find significant benefits from using a learning rate scheduler with warmup or decay for our setup. Key hyperparameters for our main ImageNet training and the single-class training setup are detailed in Table~\ref{tab:hyperparams}.

\section{Limitations}
\label{app:limitations}

Despite the strong performance of AID-VAR, several limitations remain.
First, the guidance weight $w$ is currently manually designed and kept static throughout generation. Such a fixed strategy may not be optimal across different scales, image complexities, or uncertainty levels, potentially limiting the flexibility and robustness of the guidance process.

Moreover, although AID-VAR effectively mitigates error accumulation in autoregressive image generation, extending the framework to video generation remains challenging. In autoregressive video models, errors accumulate not only spatially but also temporally across frames, often resulting in semantic drift and temporal inconsistency during long-duration synthesis. The current AID-VAR framework has not yet been validated in such temporally coherent settings, and its computational overhead may further increase when handling long video sequences.

\section{Model architectures}
\label{app: model architectures}

\subsection{Guidance injector}
The Guidance Injector is a lightweight Transformer encoder designed to be computationally efficient. It consists of 2 Transformer blocks with 8 attention heads each. The embedding dimension (\texttt{PLANNER\_DIM}) is matched to the VAR model's depth, specifically \texttt{VAR\_DEPTH * 64} (\textit{e.g.}, 1536 for VAR-d24). Within each block, the feed-forward network (FFN) has a hidden dimension of $\min(\texttt{embed\_dim} \times 2, 512)$ and uses the GELU activation function. This means the MLP ratio is approximately 0.5, 0.4, and 0.33 for our d16, d20, and d24 models, respectively. We employ a custom \texttt{SafeLayerNorm} for normalization and scale the residual connections by a factor of 0.1 to stabilize training. Sinusoidal positional encodings are added to the input features.

\subsection{Discriminator}
Our discriminator follows the Projected Discriminator architecture. We use a pre-trained DINO ViT-S/16 model \citep{caron2021emerging} as the frozen feature backbone. The trainable part consists of shallow classification heads (``DiscHead''). Each head is composed of a main block and a classification layer. The main block includes a 1x1 convolution followed by a ``ResidualBlock'' containing a 9x9 convolution. The classification layer is a 1x1 ``SpectralConv1d'' layer. During training, the entire DINO backbone remains frozen, and only the parameters of these heads are updated, ensuring efficiency and stability.

\section{Broader impact statement}
\label{app:impact}

\textbf{Positive impacts:} The AID-VAR framework presents a practical and resource-efficient paradigm for improving existing large-scale generative models. This can have positive impacts in various domains. In creative fields, it can help artists and designers generate higher-quality, more coherent images, serving as a more reliable tool for concept art and automated content creation. In scientific applications, such as medical or satellite imaging, our method's ability to reduce structural artifacts could lead to more robust and reliable image reconstruction and enhancement, aiding in diagnosis and analysis. More broadly, our work contributes to the field of trustworthy AI by providing a method to diagnose and correct flaws in foundational models post-hoc.

\textbf{Potential risks:} As with any powerful generative technology, there are potential risks. The ability to generate more realistic and coherent images could be misused for creating high-quality ``deepfakes'' or other forms of synthetic media for malicious purposes, such as spreading misinformation or creating fraudulent content. The increased quality might make such fakes harder to detect. Furthermore, while we aim to correct errors, the adversarial training process could inadvertently amplify existing biases present in the training data if not carefully monitored. The additional training step, though efficient, still contributes to the overall energy consumption and carbon footprint of developing large AI models.

\textbf{Mitigation strategies:} To mitigate these risks, we support the concurrent development of robust detection methods for synthetic media. Techniques like digital watermarking can be integrated into the generative process. We advocate for responsible use policies and clear labeling of synthetic content when deployed in public-facing applications. To address bias, it is crucial to use diverse and carefully curated datasets and to continuously evaluate model outputs for fairness across different demographic groups. Finally, we encourage research into more energy-efficient training and optimization techniques to reduce the environmental impact.


\end{document}